\documentclass{article} 
\usepackage{iclr2027_conference,times}
\usepackage{etoolbox}
\makeatletter
\patchcmd{\@maketitle}
  {\lhead{Published as a conference paper at ICLR 2027}}
  {\lhead{Preprint version}}
  {}
  {\PackageError{main}{Could not patch the ICLR preprint header}{}}
\makeatother

\usepackage{amsmath,amsfonts,bm}

\def\eqref#1{equation~\ref{#1}}

\def\1{\bm{1}}

\DeclareMathAlphabet{\mathsfit}{\encodingdefault}{\sfdefault}{m}{sl}
\SetMathAlphabet{\mathsfit}{bold}{\encodingdefault}{\sfdefault}{bx}{n}

\usepackage{xspace}
\usepackage{graphicx}
\usepackage{subcaption}
\usepackage{booktabs}
\usepackage{tabularx}
\usepackage{multirow}
\usepackage{makecell}
\usepackage[table]{xcolor}
\usepackage{amssymb}
\usepackage{mathtools}
\usepackage{wrapfig}
\usepackage{pifont}
\usepackage{enumitem}
\usepackage[textsize=tiny]{todonotes}   

\newcommand{\name}{TaH2\xspace}                
\newcommand{\supervision}{lookahead depth supervision\xspace} 
\newcommand{\Supervision}{Lookahead Depth Supervision\xspace} 
\newcommand{\tah}{TaH\xspace}                     
\newcommand{\std}{Standard\xspace}                
\newcommand{\ouro}{Ouro\xspace}                   
\newcommand{\huginn}{Huginn\xspace}               

\newcommand{\maxiter}{M}                          
\newcommand{\flops}{\mathrm{FLOPs}}
\newcommand{\decflops}{\mathrm{FLOPs}_{\mathrm{dec}}}
\newcommand{\trainflops}{\mathrm{FLOPs}_{\mathrm{train}}}

\newcommand{\ttp}{\mathrm{TPP}}                   

\newcommand{\xhdr}[1]{{\noindent\bfseries #1}.}

\newcommand{\highlightblock}[1]{\par{\centering\emph{#1}\par}}

\newcounter{finding}

\definecolor{tahgray}{RGB}{242,242,242}
\definecolor{plancolor}{RGB}{160,32,240}

\usepackage{hyperref}
\usepackage{url}
\usepackage{float}
\hypersetup{
  pdftitle={Improving Test-Time Scaling with Adaptive Looped Transformers},
  pdfauthor={Yichen You, Tianyu Fu, Aosong Feng, Xingtai Lv, Xuefei Ning, Ning Ding, Yu Wang},
  hidelinks
}

\title{Improving Test-Time Scaling with Adaptive Looped Transformers}

\author{%
    \textbf{Yichen You}$^{*1}$,
    \textbf{Tianyu Fu}$^{*1}$,
    \textbf{Aosong Feng}$^{*2}$,
    \textbf{Xingtai Lv}$^{1}$,
    \\
    \textbf{Xuefei Ning}$^{\dagger 1}$,
    \textbf{Ning Ding}$^{\dagger 1}$,
    \textbf{Yu Wang}$^{\dagger 1}$
    \\[8pt]
    \normalfont$^{1}$Tsinghua University \quad
    $^{2}$Yale University
}

\iclrfinalcopy
\begin{document}
\raggedbottom

\maketitle
\begingroup
\renewcommand{\thefootnote}{}%
\makeatletter
\long\def\@makefntext#1{\noindent#1}%
\makeatother
\begin{NoHyper}\footnotetext{$^{*}$Equal contribution. \quad $^{\dagger}$Corresponding authors.}\end{NoHyper}%
\endgroup
\begin{abstract}
\label{sec:abstract}
Looped transformers have demonstrated promising parameter efficiency by reusing layers for latent computation.
Prior studies compare looped and non-looped models at matched parameters or per-token FLOPs. 
However, to the best of our knowledge, whether looping improves test-time scaling as outputs grow longer remains underexplored.
Through post-training looped transformers, we study the accuracy--compute slope, measured as the accuracy gain per doubling of test-time decoding FLOPs.
We find that existing looped transformers often yield steeper slopes than their non-looped baseline, yet \textit{underperform it at matched compute}.
While fixed-depth looping spends extra iterations on every token, our analysis show that many tokens do not benefit from extra iterations.
We therefore propose \name, which enables the model to focus extra iterations on the tokens that benefit from looping.
It jointly post-trains the backbone and an iteration decider through \emph{\supervision}, which uses online labels indicating whether further iteration improves the prediction.
\name improves both the efficiency and attainable accuracy of test-time scaling.
On challenging AIME benchmarks, \name improves the accuracy--compute slope by 53\% (2.74 vs.\ 1.79) over the non-looped baseline, exceeding the baseline's peak accuracy by about 3.4 points at matched test-time compute.
As the maximum iteration depth increases, existing looped models largely plateau, while \name's gain over the non-looped baseline continues to grow from +2.8 points at depth 2 to +3.9 points at depth 8.
\end{abstract}
\begin{figure}[H]
    \centering
    \begin{subfigure}[t]{0.48\linewidth}
        \centering
        \includegraphics[width=\linewidth]{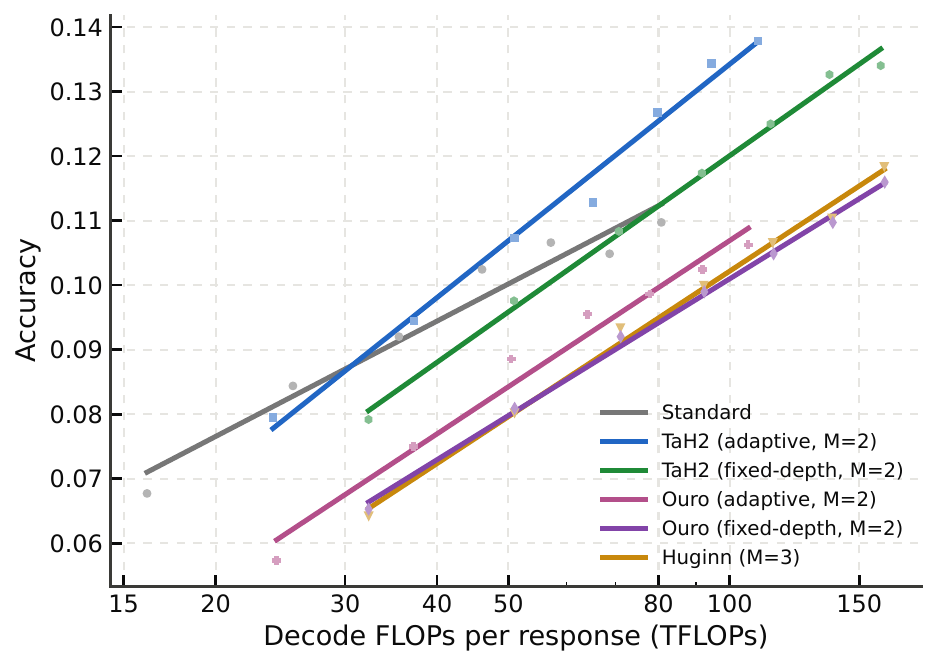}
        \caption{Test-time scaling}
        \label{fig:teaser/tts}
        \label{fig:tts_flops}
    \end{subfigure}\hfill
    \begin{subfigure}[t]{0.48\linewidth}
        \centering
        \includegraphics[width=\linewidth]{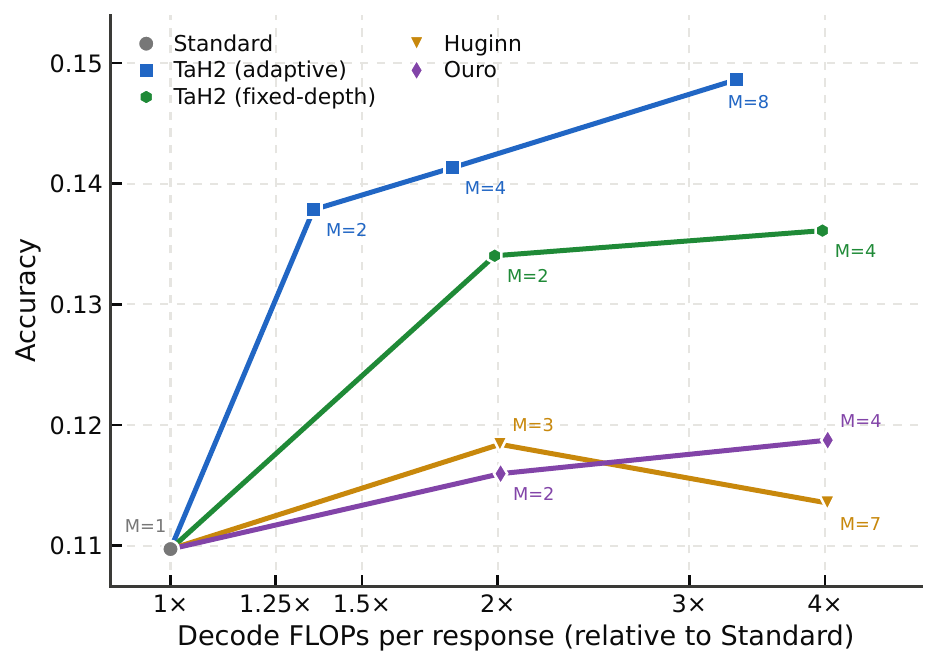}
        \caption{Iteration-depth scaling}
        \label{fig:teaser/depth}
        \label{fig:acc_depth}
    \end{subfigure}
    \caption{\name improves both test-time and iteration-depth scaling.
    Mean AIME24--26 accuracy over 32 samples for 1.7B models post-trained from the same checkpoint.
    (a)~Output-token cutoffs sweep 4K--16K; lines are fitted trends.
    (b)~Accuracy at a 16K cutoff as the depth ceiling $\maxiter$ grows.}
    \label{fig:teaser}
\end{figure}

\newpage

\begin{figure}[!t]
    \centering
    \includegraphics[width=\linewidth]{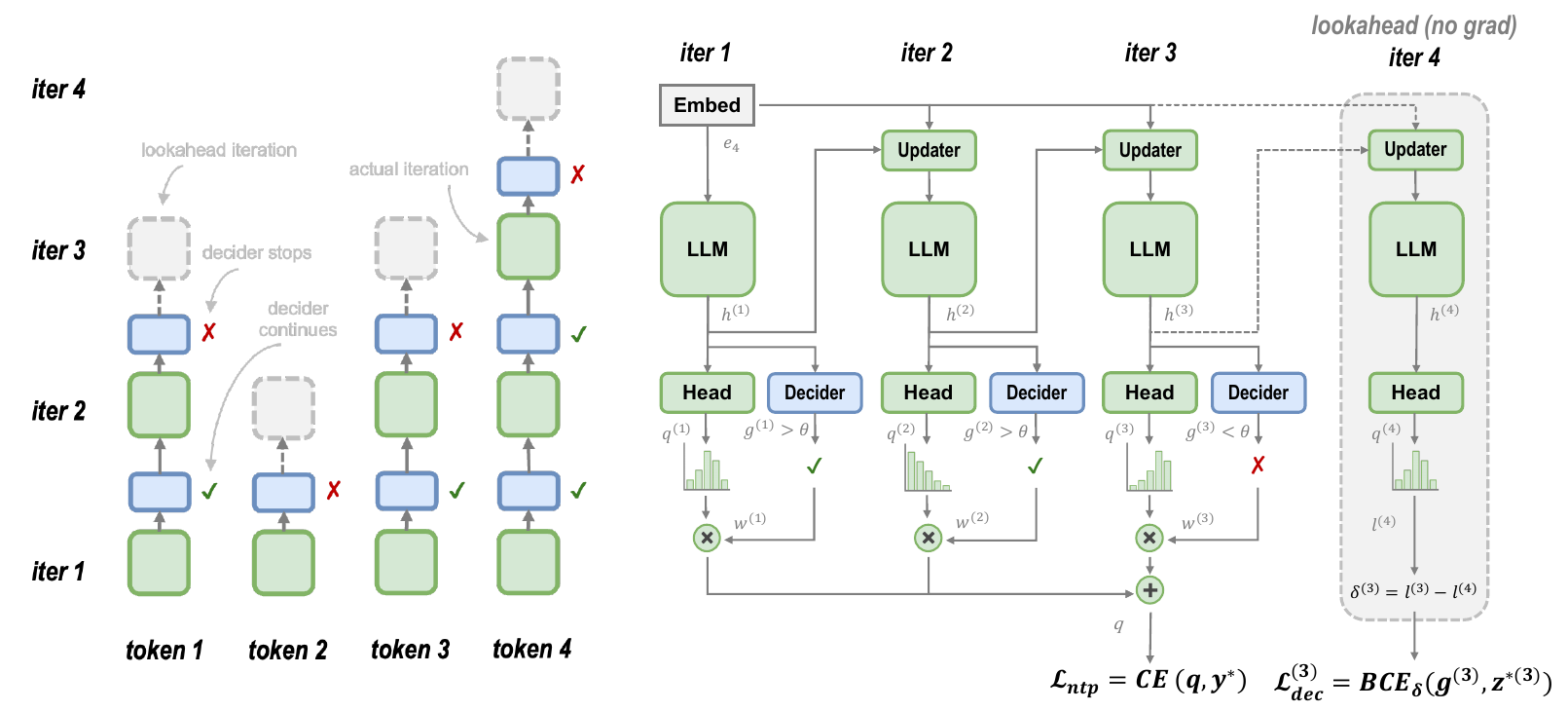}
    \caption{\name's architecture and training scheme.
    Left: the decider decides whether to continue after each iteration.
    Right: the updater reinjects token embeddings at each iteration, and stopping probabilities weight per-iteration predictions. All modules share parameters across iterations.
    The decider receives online depth supervision after each iteration; the no-gradient lookahead supplies the target at the stopping point and is omitted at inference.}
    \label{fig:arch}
\end{figure}

\section{Introduction}
\label{sec:intro}

Test-time scaling improves language model reasoning by spending additional inference compute~\citep{jaech2024openai-o1,guo2025deepseek-r1,snell2024scaling}.
This computation can support longer chains of thought in token space~\citep{muennighoff2025s1}, or repeated applications of shared layers in latent space, as in looped transformers~\citep{dehghani2018universal,geiping2025scaling,zhu2025ouro}.
Parameter sharing makes additional depth possible without increasing model size, but each iteration still incurs decoding cost.
Prior scaling studies compare looped and non-looped models at matched parameter counts or per-token FLOPs~\citep{prairie2026parcae,schwethelm2026isodepth,wang2026smelt}.
These comparisons do not establish whether looping improves accuracy--compute scaling as output-token budgets increase.
We study this question through post-training, a practical way to introduce recurrence into pretrained LLMs without training a looped model from scratch~\citep{mcleish2025retrofit,chen2025dnd,fu2025tah}.

Using the same pretrained checkpoint and post-training data, we compare \std (the non-looped baseline) with two principal looped architectures: full-stack recurrence in \ouro (at fixed depth or with its adaptive exit gate), and middle-block recurrence in \huginn~\citep{zhu2025ouro,geiping2025scaling,huang2026loopedmodels}.
By varying the output-token cutoff up to 16K at test time, we measure the scaling slope as the accuracy gain per doubling of decoding FLOPs.
In this post-training setting, fixed-depth and adaptive \ouro (maximum iteration depth $\maxiter=2$) and \huginn ($\maxiter=3$) yield steeper scaling slopes than \std ($2.12$, $2.27$ and $2.26$ versus $1.79$), yet remain less accurate over the overlapping compute range (Figure~\ref{fig:teaser/tts}).
This motivates our central question:
\highlightblock{How to post-train looped LLMs to outperform non-looped LLMs \\at the same test-time compute?}

Fixed-depth iteration spends extra compute on every token, although many tokens gain little or even get worse (Section~\ref{sec:loop/tts}).
\ouro's adaptive exit gate improves compute efficiency, but not enough to surpass \std.
We introduce \name to improve test-time scaling by learning which tokens benefit from additional iterations and allocating depth accordingly.
The backbone and an iteration decider are jointly post-trained through \emph{\supervision}: depth labels are derived online from changes in prediction loss, and a cost-sensitive loss supervises each depth decision (Figure~\ref{fig:arch}).
Unlike \tah's staged training with offline mismatch labels~\citep{fu2025tah}, \name supervises depth decisions using actual iteration gains measured on the current backbone throughout training.
In addition, an updater provides input injection between iterations~\citep{geiping2025scaling}, and stopping probabilities weight the predictions across executed depths~\citep{zeng2026adaptivelatentcot}.

We post-train Qwen3-Base models~\citep{yang2025qwen3} on general-domain data and evaluate them on math, code, QA and tool use benchmarks.
On AIME24--26 at 1.7B, \name improves the accuracy--compute slope by $53\%$ over the non-looped baseline ($2.74$ vs.\ $1.79$ points per doubling of decoding FLOPs).
With evaluation extended to 32K tokens, \name exceeds \std's peak accuracy by about $3.4$ points at matched decoding FLOPs (Figure~\ref{fig:acc_tts}).
Existing looped models largely plateau as iteration depth increases, while \name's gain over the non-looped baseline grows from $+2.8$ points at $\maxiter=2$ to $+3.9$ points at $\maxiter=8$ (Figure~\ref{fig:teaser/depth}).
\name's gains persist at larger scales (4B and 8B) and generalise beyond math to code, QA and tool use (Table~\ref{tab:size}).

We summarise our contributions as follows.
\begin{itemize}[leftmargin=20pt,topsep=3pt,itemsep=3pt,parsep=0pt]
    \item \xhdr{Test-Time Scaling of Looped Models}
    We study how looping changes accuracy--compute scaling and find that, although post-trained looped models often yield steeper slopes, they underperform \std at matched test-time compute.
    \item \xhdr{Adaptive Looped Post-training}
    We propose \name, which jointly post-trains the backbone and an iteration decider, directly supervising depth decisions with online labels indicating whether further iteration improves prediction.
    \item \xhdr{Improved Test-Time and Depth Scaling}
    On challenging AIME benchmarks, \name yields a steeper accuracy--compute slope than \std and achieves about $3.4$ points higher accuracy at matched compute when \std plateaus.
    As the maximum iteration depth increases from $2$ to $8$, its gain over \std grows from $+2.8$ points to $+3.9$ points.
\end{itemize}

\section{Related Work}
\label{sec:related_work}

\xhdr{Test-time scaling}
Test-time compute can be increased by producing more tokens or repeatedly applying a shared block in latent space.
Token-based methods generate longer chains of thought~\citep{jaech2024openai-o1,guo2025deepseek-r1}, with reasoning length controlled through test-time budget forcing or RL~\citep{muennighoff2025s1,aggarwal2025l1}.
They also explore multiple reasoning paths through repeated sampling~\citep{brown2024monkeys} or tree search~\citep{snell2024scaling,wu2024inference}.
In latent space, latent optimization replaces intermediate text with continuous representations~\citep{hao2024training,li2025implicit}, while looped transformers repeatedly apply shared layers before generating each token~\citep{geiping2025scaling,zhu2025ouro}.
We study how looping changes accuracy--compute scaling as output budgets increase, comparing post-trained looped and non-looped models at matched decoding FLOPs (Section~\ref{sec:loop-scaling-behavior}).

\xhdr{Looped transformers}
Looped transformers increase effective depth through parameter sharing~\citep{dehghani2018universal,yang2023looped,saunshi2025loopedTrans}.
Architectures differ in the span they repeat: \huginn loops a middle block~\citep{geiping2025scaling}, \ouro the full stack~\citep{zhu2025ouro}, and Loopies individual layers~\citep{gao2026loopies}.
Recent scaling studies examine recurrence under matched parameter counts~\citep{prairie2026parcae}, matched per-token FLOPs~\citep{schwethelm2026isodepth}, or both, as in SMELT~\citep{wang2026smelt}.
Recurrence can also be introduced into pretrained non-looped models through layer sharing~\citep{bae2024relaxed} or curriculum-based adaptation~\citep{mcleish2025retrofit}.
\name builds on this post-training setting to learn token-dependent iteration depths.

\xhdr{Adaptive computation}
Adaptive computation can operate along width, by selecting experts within layers~\citep{fedus2022switch}, or depth, through early exit~\citep{schuster2022confidentEarlyExit}, layer skipping~\citep{raposo2024mixture} or variable iteration counts~\citep{graves2016adaptive,banino2021pondernet}.
Within looped models, methods differ in how they learn token-dependent iteration depth.
MoR trains routers under capacity constraints~\citep{bae2025mor}, while pondering methods combine prediction loss with budget or confidence penalties~\citep{li2026ponderlm3,song2026adaponderlm,zeng2026adaptivelatentcot}.
Other approaches learn stopping decisions from terminal rewards~\citep{kuo2026rlhalting} or explicit labels: \tah uses offline mismatch labels in separate training stages~\citep{fu2025tah}, and \ouro's second stage supervises its gate with token-level loss improvements while freezing the backbone~\citep{zhu2025ouro}.
\name jointly post-trains the backbone and decider through \emph{\supervision}, deriving online token-level labels from measured iteration gains along decider-selected routes.
Appendix~\ref{sec:appendix/related} provides further comparisons and discusses scaling, state design and serving.

\section{Test-Time Scaling of Current Looped Transformers}
\label{sec:loop-scaling-behavior}
We formalise the recurrent computation of looped transformers and empirically show the test-time scaling behavior of existing looped transformers.

\subsection{Preliminaries}
\label{sec:loop/prelim}
\label{sec:method/prelim}
We use subscript $t$ for token position and superscript $(m)$ for iteration index.
We focus on models that repeat all $L$ Transformer layers, denoting the shared Transformer backbone by $\mathcal F_\theta$ and the embedding at token position $t$ by $\mathbf e_t$.
Let $\mathbf h_t^{(m)}$ denote the hidden state at token position $t$ after iteration $m$.
Starting from $\mathbf h_t^{(0)}=\mathbf e_t$, the basic forward pass applies the backbone at each iteration $m$ and computes next-token probabilities:
\begin{equation}
    \mathbf h_t^{(m)}=\mathcal F_\theta\!\left(\mathbf h_t^{(m-1)}\right),
    \qquad
    \mathbf q_t^{(m)}=\operatorname{softmax}\!\left(\mathbf W_{\mathrm{out}}\mathbf h_t^{(m)}\right),
    \label{eq:loop_general}
\end{equation}
Here $\mathbf W_{\mathrm{out}}$ is the output projection.
\ouro follows this recurrence~\citep{zhu2025ouro}.
\huginn repeats only a middle block, with input injection at each iteration~\citep{geiping2025scaling}.
Let $m_t\in\{1,\ldots,\maxiter\}$ denote the executed depth of token $t$, where $\maxiter$ is the maximum iteration depth.
Fixed-depth models use $m_t=\maxiter$ for all tokens; \std is the non-looped baseline with $m_t=1$.
Adaptive models choose $m_t$ per token, as \ouro does with a learned exit gate.

\subsection{Test-Time Scaling Behaviour}
\label{sec:loop/tts}

\xhdr{Setup}
We post-train \std, \ouro ($\maxiter=2$) and \huginn ($\maxiter=3$) from Qwen3-1.7B-Base on same data with a 16K context.
\huginn runs at fixed depth; \ouro is evaluated both at fixed depth and with its exit gate, trained afterwards on the frozen backbone (Appendix~\ref{sec:appendix/arch}).
We measure mean AIME24--26 accuracy with 32 samples per problem, sweeping output-token cutoffs from 4K to 16K in 2K increments.
We fit accuracy linearly against $\log_2$ of the decoding FLOPs per response (Appendix~\ref{sec:appendix/flops/decode}).
Further settings appear in Section~\ref{sec:exp/setup}.

\xhdr{Results}
Fixed-depth \ouro, adaptive \ouro and \huginn yield slopes of $2.12$, $2.27$ and $2.26$ accuracy points per compute doubling, versus $1.79$ for \std (Figure~\ref{fig:teaser/tts}), suggesting larger accuracy gains from latent computation as test-time compute increases.
Yet all three remain less accurate than \std over the overlapping compute range; the exit gate lowers \ouro's decoding cost but does not close the gap.
Recurrence has proven effective in pretraining, as in \ouro and \huginn, but our results point to a gap when it is introduced only in post-training:
\highlightblock{Post-trained looped models gain accuracy faster as test-time compute increases,\\ yet underperform the non-looped baseline at matched compute.}

\noindent
\begin{minipage}[t]{0.50\linewidth}
\xhdr{Analysis and motivation}
To examine how fixed-depth iteration affects individual tokens, we compare next-token losses after the first and final iterations on the validation set.
Figure~\ref{fig:loop_gain_distribution} shows the loss reduction from iterations $1\!\to\!2$ for \ouro and $1\!\to\!3$ for \huginn.
Both distributions concentrate near zero, and a substantial fraction of tokens obtain worse prediction loss after the additional iterations.
For \ouro and \huginn, respectively, $52.3\%$ and $33.5\%$ of tokens change by at most $10^{-3}$, while $21.7\%$ and $15.9\%$ become worse by more than this threshold.

Fixed-depth looping therefore spends additional iterations on many tokens that gain little or even get worse.
\ouro's exit gate improves efficiency but still underperforms \std at matched compute.
Its backbone is trained only at full depth, causing a train--inference mismatch under early exit.
This motivates learning token-dependent depth jointly with the backbone, supervised by measured loss reductions.
\end{minipage}\hfill
\begin{minipage}[t]{0.47\linewidth}
\centering
\captionsetup{type=figure}
\includegraphics[width=\linewidth]{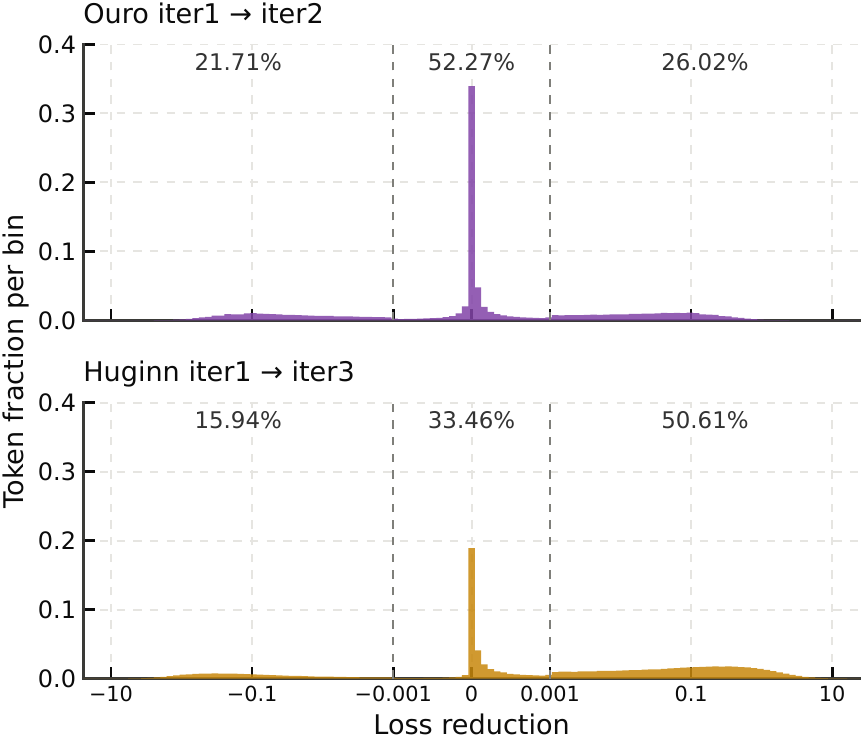}
\caption{Token-level loss reduction from the first to final iteration in \ouro and \huginn. The three regions report the fractions of tokens that become worse, change little, or improve.}
\label{fig:loop_gain_distribution}
\end{minipage}
\par

\section{\name: Post-training Adaptive Looped Models}
\label{sec:method}
Building on Section~\ref{sec:loop-scaling-behavior}, \name learns token-dependent depth through joint post-training of the backbone and decider.
We describe its architecture and training scheme below.

\subsection{Architecture}
\label{sec:method/model}
\name comprises a shared Transformer backbone, a learned input-injection updater, and a token-level iteration decider.
The updater and decider add fewer than $3\%$ parameters at every scale we study (Table~\ref{tab:params}).
Each iteration executes all $L$ backbone layers using extended duo-causal attention.

\xhdr{Extended duo-causal attention}
We extend \tah's duo-causal attention~\citep{fu2025tah} to support training-time lookahead.
At depth $m$, token $t$ attends to executed KV states at positions $s\le t$ and depths $j\le m$.
During training, stopped tokens take an additional no-gradient iteration for supervision.
Lookahead queries cannot attend to other tokens' lookahead states; all other attention follows the duo-causal rule (Appendix~\ref{sec:appendix/arch/definitions}).

\xhdr{Learned state update}
The first iteration takes the token embedding $\mathbf e_t$ as input.
Subsequent iterations use a learned input-injection updater~\citep{geiping2025scaling}, modifying Equation~\ref{eq:loop_general} to
\begin{equation}
    \mathbf h_t^{(m+1)}=\mathcal F_\theta\!\left(\mathcal U_\psi\!\left(\mathbf e_t,\mathbf h_t^{(m)}\right)\right).
    \label{eq:updater}
\end{equation}
Here $\mathcal U_\psi$ is a small normalised MLP that reinjects the input embedding while mapping the previous final-layer state back to the backbone input space (Appendix~\ref{sec:appendix/arch}).

\xhdr{Iteration decider and output}
After iteration $m<\maxiter$, a lightweight decider predicts a conditional continue probability
\begin{equation}
    g_t^{(m)}
    =\mathcal D_\phi\!\left(\mathbf e_t,\mathbf h_t^{(m)},\operatorname{TopK}(\mathbf q_t^{(m)})\right)\in(0,1).
    \label{eq:decider}
\end{equation}
With exit threshold $\tau_{\mathrm{exit}}$, token $t$ stops at the first depth where $g_t^{(m)}<\tau_{\mathrm{exit}}$ and otherwise runs to $\maxiter$.
The continue probabilities induce stopping weights over the executed depths~\citep{graves2016adaptive,zeng2026adaptivelatentcot}, with the final executed iteration absorbing the remaining mass:
\begin{equation}
    \omega_t^{(m)}=
    \begin{cases}
        (1-g_t^{(m)})\prod_{j<m}g_t^{(j)}, & m<m_t,\\[2pt]
        \prod_{j<m_t}g_t^{(j)}, & m=m_t,
    \end{cases}
    \qquad
    \mathbf q_t=\sum_{m=1}^{m_t}\omega_t^{(m)}\mathbf q_t^{(m)}.
    \label{eq:mixture}
\end{equation}
Training is on-policy with respect to depth selection: tokens follow the current decider's decisions under the same rule used at inference.
In contrast, \ouro trains at full depth before applying early exit at inference~\citep{zhu2025ouro}, while \tah trains under an oracle policy and uses learned decisions at inference~\citep{fu2025tah}.

\subsection{Training with \Supervision}
\label{sec:method/sft}
\name jointly trains the backbone, updater and decider, with the decider learning online how many iterations each token should execute.
At each iteration, \emph{\supervision} uses the measured change in prediction loss to supervise whether the token should continue (Figure~\ref{fig:arch}, right).

\xhdr{Online continuation labels}
For each supervised token $t$, let $a_t^{(m)}=\1[m_t\ge m]$ indicate whether it actually executes iteration $m$.
For tokens with $a_t^{(m)}=1$ and $m<\maxiter$, we measure iteration gains using the per-iteration predictions $\mathbf q_t^{(m)}$.
For target token $y_t^{*}$, the corresponding prediction loss and gain from another iteration are
\begin{equation}
    \ell_t^{(m)}=-\log\mathbf q_t^{(m)}[y_t^{*}],
    \qquad
    \delta_t^{(m)}=\ell_t^{(m)}-\ell_t^{(m+1)}.
    \label{eq:iteration_gain}
\end{equation}
Positive gains favour continuing; negative gains favour stopping.
For tokens that stop before $\maxiter$, a no-gradient lookahead supplies the next-iteration loss for supervision.

Let $c_t^{(m)}\in\{0,1\}$ be the target continue label after iteration $m$, with $c_t^{(0)}=1$ for all supervised tokens.
At iteration $m$, we rank positive gains among tokens with $a_t^{(m)}=1$ and $c_t^{(m-1)}=1$ and select the largest gains until they account for a fraction $\rho$ of the total.
The smallest selected gain defines the computed cutoff $\delta_{\mathrm{cut}}^{(m)}$, giving
\begin{equation}
    c_t^{(m)}=a_t^{(m)}\,c_t^{(m-1)}\,\1\!\left[\delta_t^{(m)}\ge\delta_{\mathrm{cut}}^{(m)}\right].
    \label{eq:labels}
\end{equation}
Once a token receives a stop label, its later labels remain zero.
These labels supervise the decider, whose decisions determine $a_t^{(m+1)}$.
Coverage $\rho$ retains most of the positive gain while excluding marginal improvements that can produce noisy labels.

\xhdr{Joint objective}
We combine next-token prediction loss with cost-sensitive supervision of the decider.
The joint objective is
\begin{equation}
    \mathcal L_{\mathrm{SFT}}=
    \underbrace{\frac{1}{N}\sum_t-\log\mathbf q_t[y_t^{*}]}_{\text{next-token prediction loss}}
    +\alpha_D\,
    \underbrace{\frac{1}{N}\sum_{m=1}^{\maxiter-1}\sum_{t:\,a_t^{(m)}=1}
    w_t^{(m)}\,\mathrm{BCE}\!\left(g_t^{(m)},c_t^{(m)}\right)}_{\text{cost-sensitive decider loss}},
    \label{eq:loss}
\end{equation}
where $N$ is the number of supervised tokens and $\mathbf q_t$ is the stopping-weighted mixture in Equation~\ref{eq:mixture}.
At iteration $m$, we supervise the decider on all tokens with $a_t^{(m)}=1$, using cost-sensitive weights $w_t^{(m)}$ computed from $|\delta_t^{(m)}-\delta_{\mathrm{cut}}^{(m)}|$ to strengthen supervision farther from the cutoff.
We set coverage $\rho=0.99$ and the decider-loss coefficient $\alpha_D=0.05$; further details are provided in Appendix~\ref{sec:appendix/formulation}.

\section{Experiments}
\label{sec:experiment}

\subsection{Setup}
\label{sec:exp/setup}
We summarise the key configuration here; full details are in Appendix~\ref{sec:appendix/setup}.

\xhdr{Models and baselines}
We use Qwen3-\{1.7B, 4B, 8B\}-Base~\citep{yang2025qwen3} as backbones.
We use the 1.7B model in Section~\ref{sec:exp/performance}, and the 4B and 8B models in Section~\ref{sec:exp/size}.
We compare \name with the following baselines:
(1) \emph{\std}, the single-pass model ($\maxiter=1$);
(2) \emph{\name-fixed}, a variant of \name without a decider that executes all $\maxiter$ iterations at every token position;
(3) \emph{\ouro}, which loops all $L$ layers and carries the hidden state directly across iterations~\citep{zhu2025ouro}; and
(4) \emph{\huginn}, which loops a middle span of layers, reinjects the pre-loop representation through an input adapter, and uses a fixed depth~\citep{geiping2025scaling}.
All variants at a given scale use the same Qwen3 initialization and post-training recipe.
We train and evaluate \name-fixed and \ouro at $\maxiter=2,4$, and \huginn at $\maxiter=3,7$; looping 14 of the 28 layers gives \huginn the same effective depths of $2$ and $4$ full-model passes, respectively.
Figure~\ref{fig:loss_depth} reports mean iteration depth in units of a full $L$-layer pass.

\xhdr{Training setup}
Training data combine math, code and QA prompts from AM-Qwen3-Distilled~\citep{amteam2025qwen3distilled} with tool use samples from Nemotron-Agentic-v1~\citep{nvidia2025nemotronagentic}.
For the experiments in Section~\ref{sec:exp/performance}, we use 273K prompts with responses regenerated by Qwen3-8B, the best-performing teacher in our comparison of downstream student accuracy (Appendix~\ref{sec:appendix/results/teacher}).
We train for three epochs with a 16{,}384-token context, totalling 3.4B training tokens.
Section~\ref{sec:exp/size} uses the original responses generated by Qwen3-235B-A22B and fixes the training budget at $\ttp=2$ tokens per parameter for every model size.
A validation set of 1{,}000 samples randomly drawn from the training mixture is used for loss and token-level analyses.

\xhdr{Evaluation setup}
We evaluate on math (AIME24--26, AMC23, MATH500~\citep{lightman2023let}, OlympiadBench~\citep{he2024olympiadbench}, and IMO-AnswerBench (IMO-AB)~\citep{luong2025robust}), QA (GPQA~\citep{rein2023gpqa} and SuperGPQA (SGPQA)~\citep{mapteam2025supergpqa}), code (HumanEval~\citep{chen2021codex}, MBPP~\citep{mbpp}, and LiveCodeBench~v6 (LCB)~\citep{jain2024livecodebench}), and tool use (BFCL~v3~\citep{patil2025bfcl}).
We use zero-shot CoT with temperature $0.6$, top-$p$ $0.95$, top-$k$ $20$, and a default maximum generation length of 32K tokens.
We report avg@32 on the primary math benchmarks and use fewer samples per problem on larger benchmarks (Appendix~\ref{sec:appendix/eval}).
The decider threshold is $\tau_{\mathrm{exit}}=0.5$ for \name.

\subsection{Performance}
\label{sec:exp/performance}
\label{sec:exp/depth}
\label{sec:exp/tts}

At 1.7B, \name improves accuracy across domains.
Its performance continues to improve with iteration depth, while its test-time scaling yields a steeper slope and higher accuracy at matched compute than \std.

\begin{table*}[t]
    \centering
    \caption{Accuracy (\%) of Qwen3-1.7B models across ten benchmarks.
    Olympiad: OlympiadBench; HE: HumanEval. Bold/underline indicate the best/second-best accuracy per row; subscripts show gains over \std in points. FLOPs/token is total decoding FLOPs divided by total generated tokens across benchmarks, relative to \std.}
    \label{tab:performance}
    \label{tab:appendix/all11}
    \fontsize{8}{10}\selectfont
    \setlength\tabcolsep{0pt}
    \begin{tabular}{@{}>{\raggedright\arraybackslash}p{0.11\linewidth}>{\raggedright\arraybackslash}p{0.115\linewidth}|*{7}{>{\centering\arraybackslash}p{\dimexpr0.0775\linewidth-\arrayrulewidth/5\relax}}|*{3}{>{\centering\arraybackslash\columncolor{tahgray}}p{\dimexpr0.0775\linewidth-\arrayrulewidth/5\relax}}@{}}
    \toprule
    \multirow{2}{*}{\textbf{Domain}} & \multirow{2}{*}{\textbf{Benchmark}} & \textbf{Std.} & \multicolumn{2}{c}{\textbf{\huginn}} & \multicolumn{2}{c}{\textbf{\ouro}} & \multicolumn{2}{c}{\textbf{\name-fixed}} & \multicolumn{3}{|c}{\cellcolor{tahgray}\textbf{\name}} \\
    \cmidrule(lr){3-3}\cmidrule(lr){4-5}\cmidrule(lr){6-7}\cmidrule(lr){8-9}\cmidrule(lr){10-12}
    & & $\maxiter{=}1$ & $\maxiter{=}3$ & $\maxiter{=}7$ & $\maxiter{=}2$ & $\maxiter{=}4$ & $\maxiter{=}2$ & $\maxiter{=}4$ & $\maxiter{=}2$ & $\maxiter{=}4$ & $\maxiter{=}8$ \\
    \midrule
    \multirow{6}{*}{math} & AIME24 & 11.0 & 13.9 & 10.6 & 10.5 & 14.5 & \underline{15.7} & 14.3 & \underline{15.7} & \textbf{16.3} & \textbf{16.3} \\
     & AIME25 & 13.0 & 14.2 & 14.0 & 13.9 & 13.4 & 15.0 & 15.4 & 16.5 & \underline{16.9} & \textbf{17.9} \\
     & AIME26 & 11.9 & 11.8 & 13.3 & 11.0 & 10.3 & 13.2 & 13.8 & \underline{14.3} & 14.2 & \textbf{14.5} \\
     & AMC23 & 45.3 & 45.8 & 49.0 & 45.9 & 45.9 & 51.1 & \underline{52.2} & 50.7 & 50.9 & \textbf{52.8} \\
     & MATH500 & 75.2 & 74.4 & 75.2 & 75.1 & 73.5 & 76.1 & 78.5 & 78.1 & \underline{78.6} & \textbf{79.3} \\
     & Olympiad & 40.4 & 40.6 & 40.4 & 37.9 & 37.6 & 42.1 & 44.0 & 43.8 & \underline{44.1} & \textbf{47.4} \\
    \midrule
    \multirow{2}{*}{code} & HE & 63.6 & 61.7 & 64.2 & 61.2 & 58.2 & 63.8 & 65.2 & \underline{66.8} & 65.2 & \textbf{72.6} \\
     & MBPP & 70.2 & 70.3 & 69.3 & 69.6 & 67.3 & 70.8 & 71.3 & 71.8 & \underline{72.3} & \textbf{74.4} \\
    \midrule
    QA & GPQA & 31.5 & \underline{33.7} & 33.4 & \textbf{34.0} & 31.2 & 31.9 & 32.8 & 33.1 & 33.0 & 32.0 \\
    \midrule
    tool use & BFCL & 15.3 & 15.6 & 14.8 & 14.3 & 12.1 & 14.6 & 15.5 & 15.3 & \underline{17.4} & \textbf{17.9} \\
    \midrule
    \multicolumn{2}{l|}{\textbf{Avg.}} & 37.7 & 38.2 & 38.4 & 37.3 & 36.4 & 39.4 & 40.3 & 40.6\textsubscript{/+2.9} & \underline{40.9}\textsubscript{/+3.2} & \textbf{42.5}\textsubscript{/+4.8} \\
    \multicolumn{2}{l|}{FLOPs/token} & 1.00$\times$ & 1.90$\times$ & 3.72$\times$ & 1.91$\times$ & 3.75$\times$ & 1.99$\times$ & 3.97$\times$ & 1.21$\times$ & 1.47$\times$ & 2.37$\times$ \\
    \bottomrule
    \end{tabular}
\end{table*}

\begin{figure*}[t]
    \centering
    \begin{subfigure}[t]{0.3513581\linewidth}
        \centering
        \includegraphics[width=\linewidth]{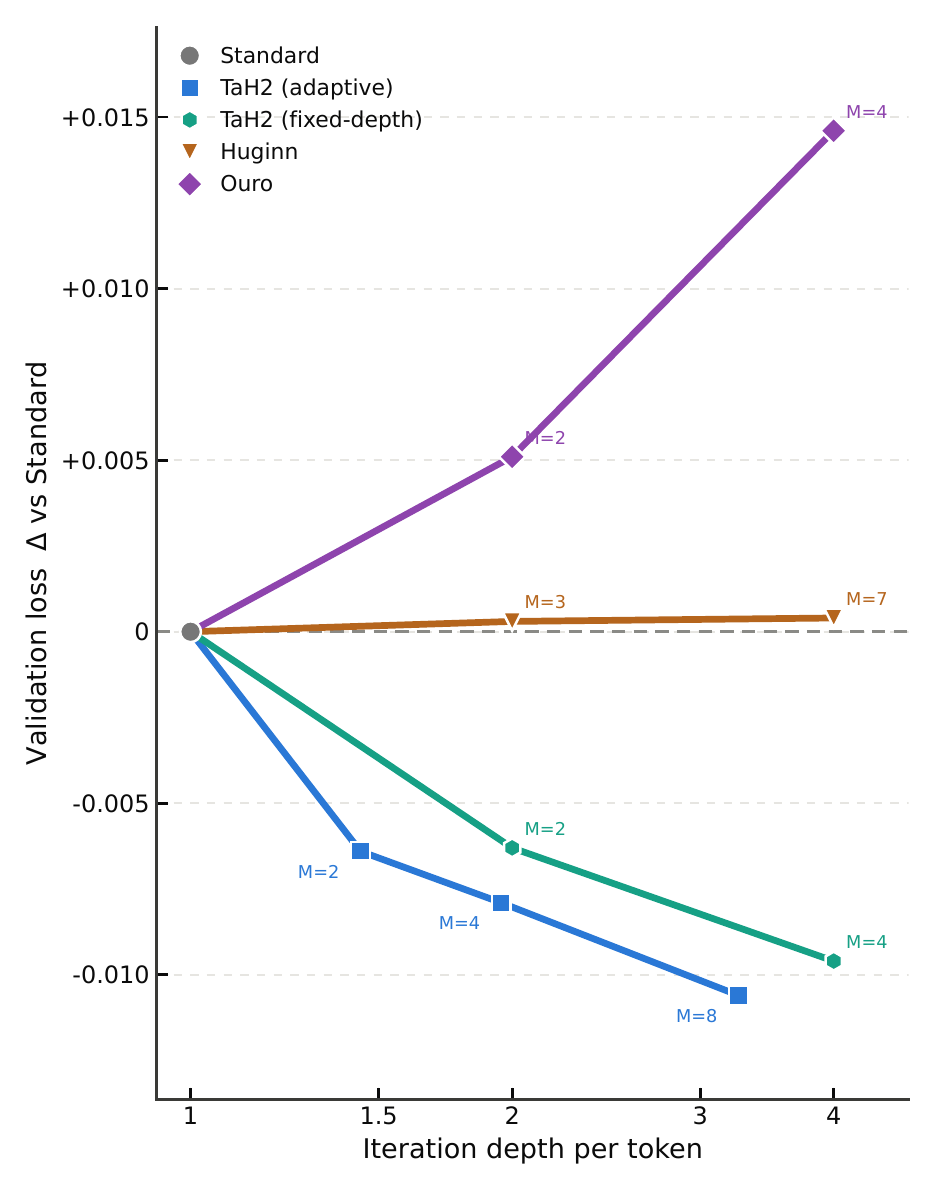}
        \caption{Final validation loss versus iteration depth.}
        \label{fig:loss_depth}
    \end{subfigure}\hfill
    \begin{subfigure}[t]{0.6286419\linewidth}
        \centering
        \includegraphics[width=\linewidth]{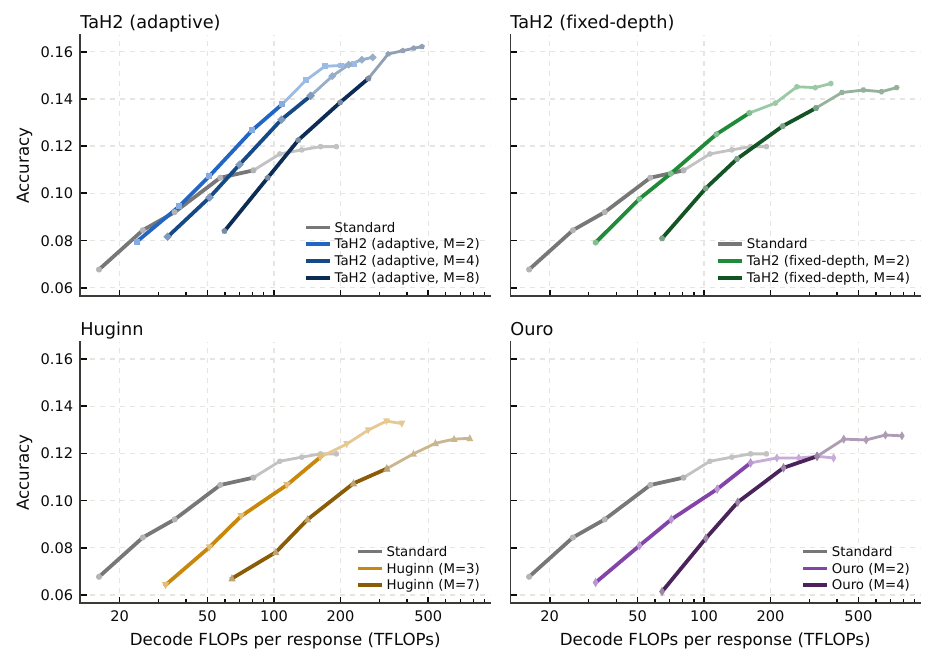}
        \caption{Mean AIME24--26 accuracy versus decoding FLOPs.}
        \label{fig:acc_tts}
    \end{subfigure}
    \caption{Depth and test-time scaling at 1.7B.
    (a)~Each marker denotes a model trained with the labelled depth ceiling $\maxiter$; lower is better.
    (b)~Dark segments show output-token cutoffs within 16K; light segments extend beyond it to 32K.}
    \label{fig:scaling}
\end{figure*}

\xhdr{Depth scaling}
Raising the depth ceiling benefits \name but not the existing looped methods.
On validation loss (Figure~\ref{fig:loss_depth}), \ouro and \huginn remain at or above \std at every ceiling, whereas both \name variants lower the loss further as $\maxiter$ grows, with adaptive \name reaching the lowest loss ($-0.0106$ versus \std at $\maxiter=8$); this advantage persists throughout training (Appendix~\ref{sec:appendix/training/curves}).
Across the ten benchmarks (Table~\ref{tab:performance}), \name's average gain over \std grows from $+2.9$ points at $\maxiter=2$ to $+4.8$ at $\maxiter=8$, whereas \huginn and \ouro stay close to \std; Figure~\ref{fig:teaser/depth} shows the same trend on AIME24--26.

\xhdr{Test-time scaling}
As decoding FLOPs increase, \name improves accuracy more efficiently than baselines and reaches a higher peak accuracy.
Within the 16K training length (Figure~\ref{fig:teaser/tts}), \name at $\maxiter=2$ gains $2.74$ points per doubling of decoding FLOPs, compared with $2.12$--$2.42$ for other looped models and $1.79$ for \std.
Extending evaluation to 32K (Figure~\ref{fig:acc_tts}), \std saturates at $12.0\%$ with $191.7$ TFLOPs per response, whereas \name reaches $15.4\%$ at the same compute, $3.4$ points higher.
Larger ceilings ($\maxiter=4,8$) raise peak accuracy further, while $\maxiter=2$ offers the best accuracy--compute trade-off among the three.
Appendix~\ref{sec:appendix/results/tts} further shows these scaling curves on each benchmark.
\name also improves parallel scaling through majority voting: AIME24--26 cons@32 reaches $27.3$--$29.3\%$ for $\maxiter=2$--$8$, versus $21.9\%$ for \std (Appendix~\ref{sec:appendix/results/cons}).

\xhdr{Runtime efficiency and real-world test-time scaling}
We serve all models with an extended Mini-SGLang engine that batches requests at different iteration depths in a shared forward pass (Appendix~\ref{sec:appendix/eval/runtime}).
Although \name adds $22\%$ decoding FLOPs per token and $30$--$34\%$ end-to-end latency relative to \std (Table~\ref{tab:efficiency}), it still achieves better test-time scaling and higher attainable accuracy in actual serving (Figure~\ref{fig:acc_time}).

\begin{table}[t]
    \newsavebox{\runtimeEfficiencyTable}
    \begin{minipage}[t]{0.60\textwidth}
    \vspace{0pt}
    \centering
    \renewcommand{\std}{Std.\xspace}
    \begin{lrbox}{\runtimeEfficiencyTable}%
    \footnotesize%
    \setlength\tabcolsep{2pt}%
    \newcommand{\rel}[1]{{\scriptsize\color{black!55}#1}}%
    \begin{tabular*}{\linewidth}{@{\extracolsep{\fill}}l cc >{\columncolor{tahgray}}c cc >{\columncolor{tahgray}}c}
    \toprule
    & \multicolumn{3}{c}{\textbf{Batch size = 1}} & \multicolumn{3}{c}{\textbf{Batch size = 4}} \\
    \cmidrule(lr){2-4}\cmidrule(lr){5-7}
    \textbf{Metric} & \textbf{\std} & \makecell{\textbf{\name-}\\\textbf{fixed}} & \cellcolor{tahgray}\textbf{\name} & \textbf{\std} & \makecell{\textbf{\name-}\\\textbf{fixed}} & \cellcolor{tahgray}\textbf{\name} \\
    \midrule
    \makecell[l]{GFLOPs/tok} & 7.04 & 14.09 & 8.59 & 7.04 & 14.09 & 8.59 \\
    \quad\rel{vs.\ \std} & \rel{1.00$\times$} & \rel{2.00$\times$} & \rel{1.22$\times$} & \rel{1.00$\times$} & \rel{2.00$\times$} & \rel{1.22$\times$} \\
    \makecell[l]{Latency(s)} & 139.8 & 328.7 & 187.0 & 220.3 & 483.3 & 285.4 \\
    \quad\rel{vs.\ \std} & \rel{1.00$\times$} & \rel{2.35$\times$} & \rel{1.34$\times$} & \rel{1.00$\times$} & \rel{2.19$\times$} & \rel{1.30$\times$} \\
    \makecell[l]{Tokens/s} & 202.3 & 89.8 & 144.3 & 507.6 & 229.9 & 376.2 \\
    \quad\rel{vs.\ \std} & \rel{1.00$\times$} & \rel{0.44$\times$} & \rel{0.71$\times$} & \rel{1.00$\times$} & \rel{0.45$\times$} & \rel{0.74$\times$} \\
    \bottomrule
    \end{tabular*}%
    \end{lrbox}
    \global\setbox\runtimeEfficiencyTable=\copy\runtimeEfficiencyTable
    \usebox{\runtimeEfficiencyTable}\par
    \captionsetup{position=bottom,width=\linewidth}
    \captionof{table}{Runtime efficiency on AIME26.
    Ratios are to \std at the same batch size; GFLOPs/token is the decode cost per generated token (Appendix~\ref{sec:appendix/flops/prelim}).}
    \label{tab:efficiency}
    \end{minipage}\hfill
    \begin{minipage}[t]{0.38\textwidth}
    \vspace{0pt}
    \centering
    \includegraphics[width=\linewidth,height=\dimexpr\ht\runtimeEfficiencyTable+\dp\runtimeEfficiencyTable\relax]{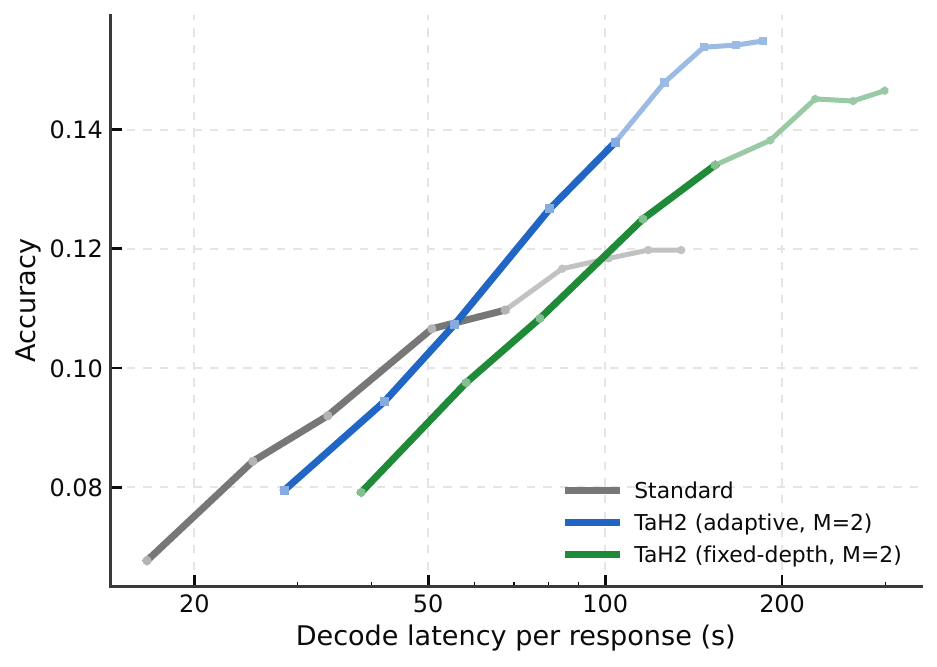}
    \captionof{figure}{Accuracy versus mean end-to-end latency.
    Lighter segments extend beyond the 16K training length.}
    \label{fig:acc_time}
    \end{minipage}
\end{table}

\subsection{Additional Scales}
\label{sec:exp/size}
We further evaluate \name ($\maxiter=2$) on 4B and 8B backbones (Table~\ref{tab:size}).
\name outperforms \std on every benchmark at both sizes, raising average accuracy by $3.2$ points at 4B and $2.4$ points at 8B.
On challenging AIME math, \name improves by up to $6.9$ points at 4B and $4.4$ points at 8B.

\begin{table}[t]
\begin{minipage}[t]{0.43\linewidth}
\vspace{0pt}
\centering
\captionof{table}{Accuracy (\%) at 4B and 8B. Subscripts give gains over \std in points.}
\label{tab:size}
\fontsize{8}{10}\selectfont
\renewcommand{\arraystretch}{1.33}
\setlength\tabcolsep{2pt}
\begin{tabular*}{\linewidth}{@{\extracolsep{\fill}}ll c >{\columncolor{tahgray}}c c >{\columncolor{tahgray}}c}
\toprule
 & & \multicolumn{2}{c}{\textbf{4B}} & \multicolumn{2}{c}{\textbf{8B}} \\
\cmidrule(lr){3-4}\cmidrule(lr){5-6}
\textbf{Domain} & \textbf{Benchmark} & \textbf{Std.} & \textbf{\name} & \textbf{Std.} & \textbf{\name} \\
\midrule
\multirow{6}{*}{math} & AIME24 & 52.0 & \textbf{57.7} & 66.4 & \textbf{70.8} \\
 & AIME25 & 40.3 & \textbf{43.1} & 52.0 & \textbf{55.9} \\
 & AIME26 & 45.5 & \textbf{52.4} & 60.9 & \textbf{63.4} \\
 & AMC23 & 88.6 & \textbf{89.2} & 93.5 & \textbf{96.6} \\
 & Olymp. & 67.0 & \textbf{68.4} & 73.3 & \textbf{74.4} \\
 & IMO-AB & 31.2 & \textbf{33.8} & 40.3 & \textbf{42.0} \\
\midrule
code & LCB & 35.4 & \textbf{37.3} & 42.1 & \textbf{44.4} \\
\midrule
QA & SGPQA & 28.6 & \textbf{30.5} & 38.0 & \textbf{39.2} \\
\midrule
tool use & BFCL & 29.2 & \textbf{33.6} & 38.0 & \textbf{39.4} \\
\midrule
\multicolumn{2}{@{}l}{Avg.} & 46.4 & \textbf{49.6}\textsubscript{/+3.2} & 56.1 & \textbf{58.5}\textsubscript{/+2.4} \\
\bottomrule
\end{tabular*}
\end{minipage}\hfill
\begin{minipage}[t]{0.55\linewidth}
\vspace{0pt}
\centering
\captionof{table}{Design choices at 1.7B. Each row varies one aspect from \name (default in \smash{\colorbox{tahgray}{gray}}).}
\label{tab:ablation}
\fontsize{8}{10}\selectfont
\renewcommand{\arraystretch}{1.225}
\setlength\tabcolsep{1pt}
\begin{tabularx}{\linewidth}{>{\raggedright\arraybackslash}X cccc}
\toprule
\textbf{Variant} & \textbf{AIME24} & \textbf{AIME25} & \textbf{AIME26} & \textbf{Avg.} \\
\midrule
\rowcolor{tahgray}
\name & \textbf{15.7} & \textbf{16.5} & \textbf{14.3} & \textbf{15.5} \\
\midrule
\multicolumn{5}{l}{\textit{Depth labels} (\name: \colorbox{tahgray}{Iteration gain})} \\
\quad Top-1 mismatch & 14.1 & 16.5 & 12.6 & 14.4{\scriptsize$/-$1.1} \\
\multicolumn{5}{l}{\textit{Decider loss weights} (\name: \colorbox{tahgray}{Cost-sensitive})} \\
\quad Uniform & 12.9 & 11.6 & 11.3 & 11.9{\scriptsize$/-$3.6} \\
\multicolumn{5}{l}{\textit{Gain coverage} (\name: \colorbox{tahgray}{$\rho=0.99$})} \\
\quad $\rho=1$ & 14.0 & 13.4 & 13.5 & 13.6{\scriptsize$/-$1.9} \\
\multicolumn{5}{l}{\textit{Training decisions} (\name: \colorbox{tahgray}{Deterministic})} \\
\quad Sampled & 14.4 & 16.3 & 12.0 & 14.2{\scriptsize$/-$1.3} \\
\midrule
\multicolumn{5}{l}{\textit{Updater} (\name: \colorbox{tahgray}{Learned})} \\
\quad Top-100 embedding & 13.3 & 14.1 & 12.7 & 13.4{\scriptsize$/-$2.1} \\
\multicolumn{5}{l}{\textit{Output} (\name: \colorbox{tahgray}{Stopping-weighted mixture})} \\
\quad Final iteration & 15.4 & 15.4 & 12.7 & 14.5{\scriptsize$/-$1.0} \\
\bottomrule
\end{tabularx}
\end{minipage}
\end{table}

\subsection{Design Choice Exploration}
\label{sec:exp/dse}
We examine the training and architectural choices of \name at 1.7B with $\maxiter=2$.
Table~\ref{tab:ablation} reports AIME24--26 accuracy under the 32K evaluation setting.

\xhdr{Training scheme}
(1) \textbf{Depth labels}.
We compare gain-based labels with top-1 mismatch labels, which label a token to continue when its top-1 prediction differs from the target, as in \tah~\citep{fu2025tah}.
Mismatch labels reduce average accuracy by $1.1$ points, as they only reflect whether the current prediction is correct, not whether further iteration actually improves it.
(2) \textbf{Decider loss weights}.
Weighting all decisions uniformly instead of by gain magnitude causes the largest drop, $3.6$ points, and lowers accuracy on all three benchmarks.
(3) \textbf{Gain coverage}.
Retaining all positive gains ($\rho=1$) reduces average accuracy by $1.9$ points, supporting the filtering of marginal gains to reduce label noise.
(4) \textbf{Training decisions}.
Sampling rather than thresholding continuation decisions during training reduces average accuracy by $1.3$ points.
This suggests that consistent token selection across training and inference helps each depth specialise.

\xhdr{Model architecture}
(1) \textbf{Updater}.
We replace the learned updater with the probability-weighted sum of the top-100 token embeddings~\citep{fu2025tah}.
This alternative reduces average accuracy by $2.1$ points, supporting learned state updates.
(2) \textbf{Output}.
Using only the final prediction reduces average accuracy by $1.0$ point, supporting the stopping-weighted mixture.

\subsection{Further Analysis}
\label{sec:exp/analysis}

\noindent
\begin{minipage}[t]{0.48\linewidth}
\vspace{0pt}
\xhdr{Decider--gain alignment}
For \name ($\maxiter=2$) at 1.7B, we group validation tokens by continue probability and measure the mean loss reduction from a second iteration (Figure~\ref{fig:gain_vs_prob}).
Tokens with probabilities near zero show negative or negligible gains, while mean gain increases with continue probability overall.
At the decision threshold of $0.5$, the corresponding loss reduction is near zero.
These results support the decider's ability to direct additional iterations towards tokens that benefit more from them.
\end{minipage}\hfill
\begin{minipage}[t]{0.48\linewidth}
\vspace{0pt}
\centering
\captionsetup{type=figure}
\includegraphics[width=\linewidth]{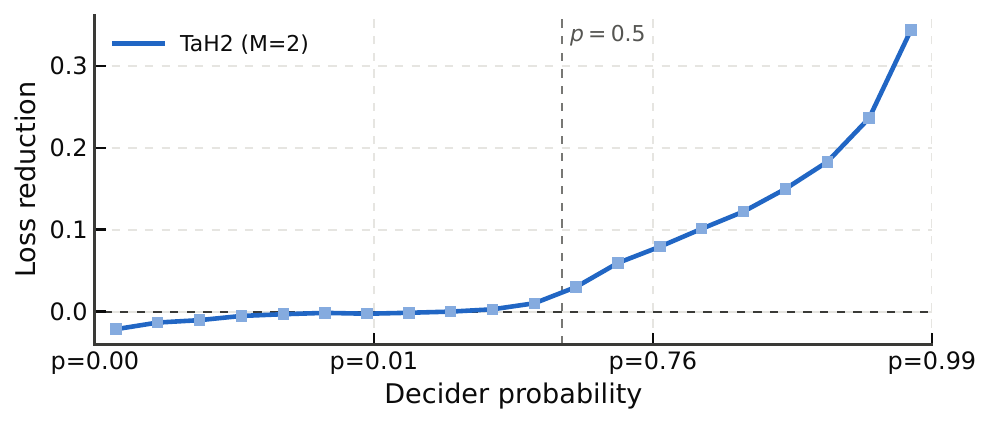}
\caption{Mean loss reduction from a second iteration versus the decider's continue probability for \name ($\maxiter=2$) on the validation set.}
\label{fig:gain_vs_prob}
\end{minipage}

\par

\par

\xhdr{Token-level depth allocation}
We visualise token-level iteration depths in sampled responses from OlympiadBench, HumanEval and GPQA (Appendix~\ref{sec:appendix/analysis/token_depth}).
In the math and code examples, mathematical expressions and final code use fewer iterations than the preceding natural-language reasoning, whereas the QA example maintains greater depth throughout.

\xhdr{Cross-iteration attention}
We examine attention across iterations in three representative heads on validation sequences (Appendix~\ref{sec:appendix/analysis/attention}).
We find that different attention heads learn distinct iteration preferences: attending primarily to first-iteration states, later-iteration states, or both.

\section{Conclusion}
\label{sec:conclusion}
In this paper, we study the test-time scaling of looped transformers introduced through post-training.
Existing looped models yield steeper accuracy--compute slopes than their non-looped baseline, yet remain less accurate at matched compute.
We therefore introduce \name, which post-trains adaptive looped transformers by jointly training the backbone and an iteration decider through \emph{\supervision}.
On AIME, \name improves the slope by $53\%$ and exceeds \std's peak accuracy by about $3.4$ points at matched compute.
Its gain over \std grows from $2.8$ to $3.9$ points as the maximum iteration depth increases from $2$ to $8$, and extends to larger models (4B and 8B) and other domains (code, QA and tool use).

\xhdr{Limitations}
(1) \name incurs more training FLOPs than standard SFT (Appendix~\ref{sec:appendix/flops/train}).
But note that post-training requires substantially less compute than pretraining, making \name a practical way to add adaptive depth to existing models.
(2) Our method is studied only under SFT; we leave its extension to on-policy distillation and reinforcement learning for future work.


\subsection*{AI use statement}
In this work, we used generative AI tools for polishing the paper text.
We have not used generative AI tools for research ideation, methodology or experimental design, method implementation, result interpretation, etc.; the remaining required-disclosure tasks are not applicable to this work.
We have reviewed all AI-assisted work: all AI-assisted text was checked by the authors against the experimental records and cited sources.

\subsection*{Ethics statement}
This study trains and evaluates language models on publicly available reasoning data and benchmarks; it does not involve human subjects or sensitive personal data.

\subsection*{Reproducibility statement}
All experiments use publicly available base models and data.
Section~\ref{sec:method} specifies the model family and objective; Appendices~\ref{sec:appendix/setup} and~\ref{sec:appendix/formulation} give the full loss, architecture and baseline definitions, FLOPs accounting, training hyper-parameters and evaluation protocol.
Training and evaluation code, configuration files, and checkpoints will be released upon publication.


\bibliography{main}
\bibliographystyle{iclr2027_conference}

\newpage
\appendix

\renewcommand{\topfraction}{0.9}
\renewcommand{\bottomfraction}{0.8}
\renewcommand{\textfraction}{0.1}
\renewcommand{\floatpagefraction}{0.8}
\setcounter{topnumber}{3}
\setcounter{totalnumber}{5}

\section{Additional Experiment Setups}
\label{sec:appendix/setup}

\subsection{Training Recipe}
\label{sec:appendix/training}

\subsubsection{Hyper-parameters}
\label{sec:appendix/training/hparams}
\begin{table}[!htbp]
    \caption{Training hyper-parameters shared by all variants at a given scale.}
    \label{tab:hparams}
    \centering
    \fontsize{8}{10}\selectfont
    \begin{tabular}{ll}
    \toprule
    \textbf{Hyper-parameter} & \textbf{Value} \\
    \midrule
    global batch (samples) & 128 \\
    sequence length & 16384 \\
    learning rate & $4\times10^{-5}$ \\
    max gradient norm & 1.0 \\
    epochs & 3 \\
    warmup ratio & 0.03 \\
    learning-rate scheduler & cosine \\
    minimum learning-rate ratio & 0.1 \\
    TPP & 2 \\
    precision & bf16\\
    coverage $\rho$ / $\alpha_D$ / threshold $\tau_{\mathrm{exit}}$ & 0.99 / 0.05 / 0.5 \\
    \bottomrule
    \end{tabular}
\end{table}

\subsubsection{Training Data}
\label{sec:appendix/training/data}
\xhdr{Main experiments}
For the 1.7B experiments (Section~\ref{sec:exp/performance}), we regenerate responses to the AM-Qwen3-Distilled and Nemotron-Agentic-v1 prompts~\citep{amteam2025qwen3distilled,nvidia2025nemotronagentic} with Qwen3-8B.
We choose this teacher because it gives the most accurate student among the teachers we compare (Appendix~\ref{sec:appendix/results/teacher}).

\xhdr{Additional scales}
For the experiments in Section~\ref{sec:exp/size}, we train the 4B and 8B backbones on the original AM-Qwen3-Distilled responses from Qwen3-235B-A22B~\citep{amteam2025qwen3distilled}, covering math, code and QA, together with tool use samples from Nemotron-Agentic-v1~\citep{nvidia2025nemotronagentic}.
We exclude samples whose combined prompt and response exceed 16{,}384 tokens.
The 4B and 8B models are trained on 8B and 16B tokens, respectively.
At each size, \name uses $\maxiter=2$ and the same training recipe as \std.

\subsection{Architecture and Baseline Details}
\label{sec:appendix/arch}
We detail the updater and decider introduced in Section~\ref{sec:method/model}, followed by the baseline implementations summarised in Table~\ref{tab:arch}.

\subsubsection{Architecture Definitions}
\label{sec:appendix/arch/definitions}
\begin{figure}[!htbp]
    \centering
    \includegraphics[width=\linewidth]{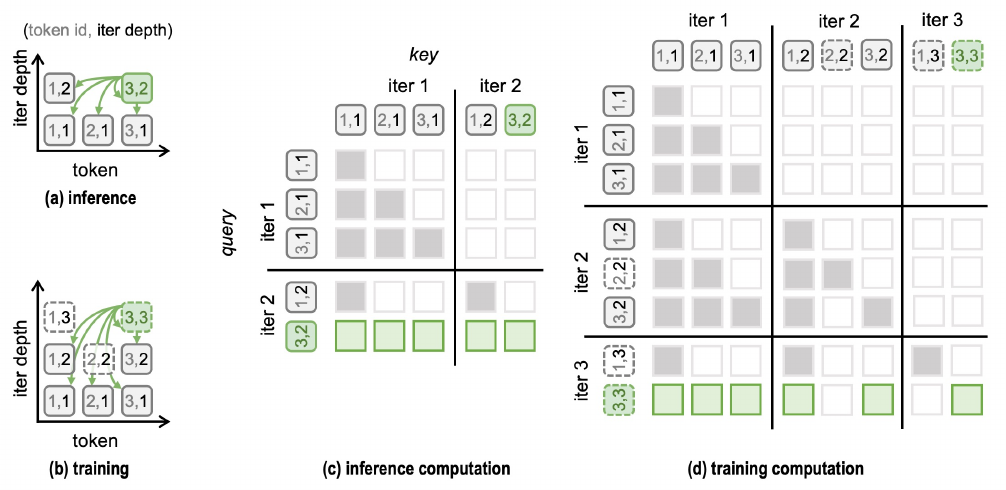}
    \caption{Extended duo-causal attention in \name.
    Each cell denotes a (token id, iteration depth) pair; green arrows and cells show the keys visible to token~3.
    (a,\,b)~Executed depths at inference and training; dashed cells are no-gradient lookahead iterations used only for depth supervision.
    (c,\,d)~The corresponding attention masks over concatenated per-iteration KV caches, where shaded cells are visible.
    Lookahead queries cannot attend to other tokens' lookahead states; all other attention follows the duo-causal rule.}
    \label{fig:duo_attention}
\end{figure}

\xhdr{Extended duo-causal attention}
Following \tah~\citep{fu2025tah}, each iteration keeps its own KV cache, and a query at token $t$ and depth $m$ attends to executed states at positions $s\le t$ and depths $j\le m$ (Figure~\ref{fig:duo_attention}).
Concatenating the per-iteration caches into one sequence turns this rule into a block-structured mask, so training and prefill process all tokens in parallel.
During training, a stopped token also computes a no-gradient lookahead iteration.
Lookahead queries cannot attend to other tokens' lookahead states, but can attend to their executed states under the duo-causal rule.
Self-attention is retained, and all other attention follows the same position and depth constraints.

\xhdr{Qwen3 MLP}
The updater and decider each use a Qwen3 SwiGLU MLP, written as $\mathcal B$:
\begin{equation}
    \mathcal B(\mathbf x)=\mathbf W_{\mathrm{down}}\big[\mathrm{SiLU}(\mathbf W_{\mathrm{gate}}\mathbf x)\odot(\mathbf W_{\mathrm{up}}\mathbf x)\big],
\end{equation}
where $\odot$ denotes elementwise multiplication.
We write $\operatorname{RN}$ for RMSNorm and $[\cdot;\cdot]$ for concatenation.
The two MLPs, $\mathcal B_{\mathrm u}$ and $\mathcal B_{\mathrm d}$, have separate weights, and each $\operatorname{RN}$ operation has its own learned scale.

\xhdr{Updater}
The updater fuses the original token embedding with the current hidden state and produces the next iteration's input:
\begin{equation}
    \begin{aligned}
    \mathbf x_{\mathrm u,t}^{(m)}
    &=\mathbf W_{\mathrm u}[\operatorname{RN}(\mathbf e_t);\operatorname{RN}(\mathbf h_t^{(m)})],\\
    \mathcal U_\psi(\mathbf e_t,\mathbf h_t^{(m)})
    &=\operatorname{RN}\!\left(\mathcal B_{\mathrm u}\!\left(\operatorname{RN}(\mathbf x_{\mathrm u,t}^{(m)})\right)\right).
    \end{aligned}
\end{equation}

\xhdr{Decider}
The decider combines the same embedding and hidden state with the largest prediction probabilities to predict whether to continue:
\begin{equation}
    \begin{aligned}
    \mathbf x_{\mathrm d,t}^{(m)}
    &=\mathbf W_{\mathrm d}[\operatorname{RN}(\mathbf e_t);\operatorname{RN}(\mathbf h_t^{(m)});\operatorname{RN}(\operatorname{TopK}(\mathbf q_t^{(m)}))],\\
    g_t^{(m)}
    &=\sigma\!\left(\mathbf W_{\mathrm{score}}\,\operatorname{RN}\!\left(\mathcal B_{\mathrm d}(\mathbf x_{\mathrm d,t}^{(m)})\right)\right).
    \end{aligned}
\end{equation}
Here $\sigma$ is the sigmoid function, and $\operatorname{TopK}$ returns probability values in descending order (2,048 values for the 1.7B model).
All projections are bias-free, and both modules share their parameters across iterations.

\xhdr{Baseline implementations}
All baselines are post-trained from the same Qwen3-1.7B-Base checkpoint using the same data, sample order, training recipe and evaluation protocol as \name.
Our \ouro and \huginn implementations adapt their architectures to this setting rather than reproduce the released checkpoints; neither uses \name's updater, decider or online supervision.

\xhdr{\name-fixed}
This variant retains \name's updater, replaces extended duo-causal attention with causal attention and removes the decider.
Every token executes all $\maxiter$ iterations.
The model is trained with next-token prediction loss on the uniformly averaged distribution, $\mathbf q_t=\frac{1}{\maxiter}\sum_{m=1}^{\maxiter}\mathbf q_t^{(m)}$.

\xhdr{\ouro}
Our implementation loops the entire backbone, passing the hidden state unchanged to the next iteration as in Equation~\ref{eq:loop_general}, and follows the two-stage training of \citet{zhu2025ouro}.
To avoid gate collapse in post-training, where every token exits after the first iteration, we simplify Stage I by removing the exit gate and uniformly averaging the per-iteration losses, $\mathcal L_{\mathrm{Ouro}}=\frac{1}{N}\sum_t\frac{1}{\maxiter}\sum_{m=1}^{\maxiter}\ell_t^{(m)}$, with every token executing $\maxiter$ iterations.
Unless otherwise stated, \ouro results use this fixed-depth model.

Stage II follows the original gate training~\citep[Section~3.4]{zhu2025ouro}.
From the Stage I checkpoint at $\maxiter=2$, we freeze the backbone and train a linear exit gate $\lambda_t^{(m)}=\sigma(\mathbf w_{\mathrm{exit}}^{\top}\operatorname{sg}[\mathbf h_t^{(m)}])$, where $\operatorname{sg}[\cdot]$ stops gradients.
With all iterations executed, the gate is supervised by the soft target $\tilde c_t^{(m)}=\sigma\big(k(I_t^{(m)}-\gamma)\big)$, where $I_t^{(m)}=\max\big(0,\operatorname{sg}[\ell_t^{(m)}-\ell_t^{(m+1)}]\big)$:
\begin{equation}
    \mathcal L_{\mathrm{gate}}=\frac{1}{\maxiter}\sum_{m=1}^{\maxiter-1}\frac{1}{N}\sum_t\mathrm{BCE}\!\left(1-\lambda_t^{(m)},\ \tilde c_t^{(m)}\right),
\end{equation}
using the original $k=50$ and $\gamma=0.005$.
Training settings follow our main post-training setup (Appendix~\ref{sec:appendix/training/hparams}), except for a learning rate of $10^{-4}$ and a single epoch.
At inference, the original Q-exit rule stops each token at the first depth whose cumulative exit probability reaches $q=0.5$.
Prefill executes all iterations; during decoding, an early-exiting token skips the remaining iterations and reuses its last executed KV entries for deeper ones~\citep{zhu2025ouro}.

\xhdr{\huginn}
Following the middle-block recurrence of \citet{geiping2025scaling}, our 28-layer backbone repeats layers 8--21 (14 layers).
Layers 1--7 are evaluated once to produce $\mathbf z_t$.
After iteration $m$, an input adapter combines $\mathbf z_t$ with the middle-block output $\mathbf r_t^{(m)}$ to form the next input:
\begin{equation}
    \mathbf u_t^{(m+1)}=\mathbf a\odot\mathbf r_t^{(m)}+\mathbf v\odot\mathbf W_{\mathrm{in}}\mathbf z_t,\qquad
    \mathbf v=\mathrm{softplus}(\mathbf b_{\mathrm{in}}),\quad \mathbf a=\exp\big(-\mathbf v\odot\exp(\mathbf b_{\mathrm{ret}})\big),
\end{equation}
where $\mathbf b_{\mathrm{in}}$ and $\mathbf b_{\mathrm{ret}}$ are learned bias vectors, $\mathbf a$ and $\mathbf v$ control state retention and input injection.
The adapter adds $d^2+2d=4.20$M parameters for hidden dimension $d=2048$ ($0.24\%$).
Every token executes $\maxiter$ iterations of the middle block, after which layers 22--28 and the LM head produce the prediction used for next-token cross-entropy.
This gives an effective depth of $14+14\maxiter$ layers: \huginn at $\maxiter=3,7$ matches the 56 and 112 layers executed by \ouro at $\maxiter=2,4$, respectively.

\subsubsection{Architecture Summary and Parameter Counts}
\label{sec:appendix/params}
Table~\ref{tab:arch} summarises the architectures; Table~\ref{tab:params} separates the updater and decider parameters at each scale. Together, they account for less than $3\%$ of total parameters.

\begin{table}[!htbp]
    \caption{Architectures used in the 1.7B comparison. Added parameters are reported as counts and fractions of total model parameters.}
    \label{tab:arch}
    \centering
    \fontsize{8}{10}\selectfont
    \setlength\tabcolsep{5pt}
    \begin{tabular}{lllll}
    \toprule
    \textbf{Model} & \textbf{Loop scope} & \textbf{Depth allocation} & \textbf{Added module} & \textbf{Added params} \\
    \midrule
    \std & none ($\maxiter=1$) & -- & -- & 0 \\
    \rowcolor{tahgray}
    \name & all $L$ layers & per-token adaptive & updater + decider & 46.2M (2.6\%) \\
    \name-fixed & all $L$ layers & fixed & updater & 21.0M (1.2\%) \\
    \ouro (fixed-depth) & all $L$ layers & fixed & -- & 0 \\
    \ouro (adaptive) & all $L$ layers & per-token adaptive & exit gate & 2.0K ($<$0.001\%) \\
    \huginn & layers 8--21 & fixed & input adapter & 4.2M (0.24\%) \\
    \bottomrule
    \end{tabular}
\end{table}

\begin{table}[!htbp]
    \caption{Updater and decider parameters across backbone scales. Counts include projections, MLPs and RMSNorm scales and are rounded independently. Percentages are relative to total model parameters.}
    \label{tab:params}
    \centering
    \fontsize{8}{10}\selectfont
    \begin{tabular}{lrrrrr}
    \toprule
    \textbf{Scale} & \textbf{Backbone} & \textbf{Updater} & \textbf{Decider} & \textbf{Total added} & \textbf{Added (\%)} \\
    \midrule
    1.7B & 1720.57M & 20.98M & 25.18M & 46.16M & 2.61\% \\
    4B & 4022.47M & 32.78M & 39.33M & 72.11M & 1.76\% \\
    8B & 8190.74M & 83.90M & 100.68M & 184.59M & 2.20\% \\
    \bottomrule
    \end{tabular}
\end{table}

\subsection{Evaluation Protocol}
\label{sec:appendix/eval}

\begin{table}[!htbp]
    \caption{Benchmarks used in this paper. The default output-token limit is 32K.}
    \label{tab:benchmarks}
    \centering
    \fontsize{8}{10}\selectfont
    \setlength\tabcolsep{2pt}
    \begin{tabular}{llrll}
    \toprule
    \textbf{Benchmark} & \textbf{Domain} & \textbf{\#Problems} & \textbf{Metric} & \textbf{Subset} \\
    \midrule
    AIME24 / AIME25 / AIME26 & math & 30 / 30 / 30 & avg@32 & -- \\
    AMC23 & math & 40 & avg@32 & -- \\
    MATH500~\citep{lightman2023let} & math & 500 & avg@4 & -- \\
    OlympiadBench~\citep{he2024olympiadbench} & math & 675 & avg@4 & -- \\
    IMO-AnswerBench & math & 400 & avg@4 & -- \\
    GPQA~\citep{rein2023gpqa} & QA & 198 & avg@8 & Diamond \\
    SuperGPQA & QA & 7{,}050 & avg@4 & Hard \\
    HumanEval~\citep{chen2021codex} & code & 164 & avg@8 & -- \\
    MBPP~\citep{mbpp} & code & 378 & avg@8 & -- \\
    LiveCodeBench~v6~\citep{jain2024livecodebench} & code & 175 & avg@4 & -- \\
    BFCL~v3~\citep{patil2025bfcl} & tool use & 800 & avg@1(greedy) & Multi-turn \\
    \bottomrule
    \end{tabular}
\end{table}
Validation loss and the token-level analyses of Section~\ref{sec:exp/analysis} use a fixed validation set of 1{,}000 samples randomly drawn from the training mixture and excluded from training, scored under teacher forcing.
Test-time scaling curves sweep the output-token cutoff and re-evaluate truncated responses; each figure specifies its cutoff range.
For the 4K--16K curves in Figure~\ref{fig:teaser/tts}, linear fits of accuracy against $\log_2$ decoding FLOPs yield $R^2=0.997$ for \ouro, $0.994$ for \huginn and $0.955$ for \std.
We estimate decoding FLOPs from the serving engine's depth and token records as described in Appendix~\ref{sec:appendix/flops}.

\xhdr{Runtime measurement}
\label{sec:appendix/eval/runtime}
We extend Mini-SGLang~\citep{xu2025minisglang} to batch requests at different iteration depths in a shared forward pass, following continuous depth batching~\citep{schwethelm2026cdb}.
Once a request completes the iterations for its current token, it proceeds to decode the next token without waiting for other requests to finish their iterations.
We evaluate the 30 AIME26 problems with a 32K output-token limit on a single A800-80GB GPU, using bfloat16 and temperature $0.6$; batch sizes 1 and 4 use one and four samples per problem, respectively.
Table~\ref{tab:efficiency} reports decoding GFLOPs per generated token, computed from the per-call costs in Appendix~\ref{sec:appendix/flops/prelim}, together with mean end-to-end latency and aggregate output-token throughput.
All results use the same evaluation seed.

\subsection{Compute Accounting}
\label{sec:appendix/flops}
This section defines the decoding FLOPs reported throughout the paper and the corresponding training cost.
Section~\ref{sec:appendix/flops/prelim} fixes the conventions and per-call costs, and Sections~\ref{sec:appendix/flops/decode} and~\ref{sec:appendix/flops/train} give compact forms of decoding and training FLOPs before expanding each term.

\subsubsection{Preliminaries}
\label{sec:appendix/flops/prelim}
\xhdr{Conventions}
We count model matrix-multiplication FLOPs, with a multiply-add as two operations.
Elementwise operations (normalization, activations, softmax, top-$k$ selection and the stopping-weighted mixture), sampling, memory traffic and serving overhead are excluded.
A linear map from $a$ to $b$ features therefore costs $2ab$ FLOPs per token.

\xhdr{Notation}
The $L$-layer backbone has hidden width $d$, total key--value width $d_{\mathrm{kv}}$ and MLP width $d_{\mathrm{ff}}$, and $V$ is the vocabulary size.
For the Qwen3 models used here, the query width $n_{\mathrm h}d_{\mathrm{head}}$ equals $d$.
The \name updater and decider have MLP widths $d_{\mathrm u}$ and $d_{\mathrm d}$, and the decider receives the $k$ largest prediction probabilities (Appendix~\ref{sec:appendix/arch/definitions}).
Token $t$ executes $m_t\le\maxiter$ iterations.

\xhdr{Per-call costs}
One backbone pass, one LM-head call, and one call of each \name module cost
\begin{equation}
    \begin{aligned}
    \flops_{\mathrm{bb}}&=L\big[4d(d+d_{\mathrm{kv}})+6d\,d_{\mathrm{ff}}\big],
    &\flops_{\mathrm{head}}&=2dV,\\
    \flops_{\mathrm{upd}}&=4d^2+6d\,d_{\mathrm u},
    &\flops_{\mathrm{decider}}&=2d(2d+k)+6d\,d_{\mathrm d}+2d.
    \end{aligned}
    \label{eq:percall}
\end{equation}
In $\flops_{\mathrm{bb}}$, $4d(d+d_{\mathrm{kv}})$ covers the query, key, value and output projections and $6d\,d_{\mathrm{ff}}$ the three SwiGLU matrices.
The updater applies $\mathbf W_{\mathrm u}$ ($2d\!\to\!d$) and $\mathcal B_{\mathrm u}$, and the decider applies $\mathbf W_{\mathrm d}$ ($2d+k\!\to\!d$), $\mathcal B_{\mathrm d}$ and $\mathbf W_{\mathrm{score}}$ ($d\!\to\!1$).
Attention cost depends on context length: a query that attends to $S$ keys adds $\flops_{\mathrm{attn}}(S)=4LdS$ for the $QK^{\top}$ and attention-weighted value products.
Table~\ref{tab:percall} lists all per-call costs for the 1.7B model; the backbone pass dominates, and the updater and decider each cost less than $2\%$ of it.

\begin{table}[!htbp]
    \caption{Per-call matrix-multiplication FLOPs for Qwen3-1.7B ($L=28$, $d=2048$, $d_{\mathrm{kv}}=1024$, $d_{\mathrm{ff}}=6144$, $V=151{,}936$, $k=d_{\mathrm u}=d_{\mathrm d}=2048$).}
    \label{tab:percall}
    \centering
    \fontsize{8}{10}\selectfont
    \setlength\tabcolsep{6pt}
    \begin{tabular}{lll}
    \toprule
    \textbf{Component} & \textbf{FLOPs per call} & \textbf{1.7B (GFLOPs)} \\
    \midrule
    Backbone pass & $L[4d(d+d_{\mathrm{kv}})+6d\,d_{\mathrm{ff}}]$ & 2.819 \\
    LM head & $2dV$ & 0.622 \\
    Attention, per visible key & $4Ld$ & $2.29\times10^{-4}$ \\
    \name updater & $4d^2+6d\,d_{\mathrm u}$ & 0.042 \\
    \name decider & $2d(2d+k)+6d\,d_{\mathrm d}+2d$ & 0.050 \\
    \huginn input adapter & $2d^2$ & 0.008 \\
    \bottomrule
    \end{tabular}
\end{table}

\subsubsection{Decoding FLOPs}
\label{sec:appendix/flops/decode}
\xhdr{Compact form}
For \name, summing over token positions processed during decoding gives
\begin{equation}
    \begin{aligned}
    \decflops=\sum_{t}\Big[&\sum_{m=1}^{m_t}\flops_{\mathrm{pass}}(t,m)
    +(m_t-1)\,\flops_{\mathrm{upd}}\\
    &+\min(m_t,\maxiter-1)\,\flops_{\mathrm{decider}}\Big],
    \end{aligned}
    \label{eq:decflops}
\end{equation}
where $\flops_{\mathrm{pass}}(t,m)$ includes the backbone, attention and LM head.
The updater prepares each iteration after the first, and the decider is not called after the final permitted iteration.

\xhdr{Positions and visible keys}
Consider one response with a $P$-token prompt.
Its first output token is sampled from the prefill, which we exclude, so decoding processes positions $t=1,\dots,N$, where $N$ is the number of output tokens minus one.
Prompt position $j$ keeps KV streams for $r_j$ iterations.
Under extended duo-causal attention, a query at position $t$ and depth $m$ sees every KV entry from earlier positions at depths up to $m$, together with its own entries from depths $1$ to $m$:
\begin{equation}
    S_t^{(m)}=\sum_{j=1}^{P}\min(r_j,m)+\sum_{u<t}\min(m_u,m)+m.
    \label{eq:duo_keys}
\end{equation}

\xhdr{Expanded form}
Each pass applies the LM head, whose per-depth predictions feed the decider and the stopping-weighted mixture, so $\flops_{\mathrm{pass}}(t,m)=\flops_{\mathrm{bb}}+\flops_{\mathrm{head}}+\flops_{\mathrm{attn}}(S_t^{(m)})$.
Substituting into Equation~\ref{eq:decflops} gives
\begin{equation}
    \begin{aligned}
    \decflops^{\text{\name}}
    =\sum_{t=1}^{N}\Big[&m_t\big(\flops_{\mathrm{bb}}+\flops_{\mathrm{head}}\big)
    +4Ld\sum_{m=1}^{m_t}S_t^{(m)}\\
    &+(m_t-1)\,\flops_{\mathrm{upd}}
    +\min(m_t,\maxiter-1)\,\flops_{\mathrm{decider}}\Big].
    \end{aligned}
    \label{eq:decflops_expanded}
\end{equation}

\xhdr{Baselines}
Fixed-depth models execute all $\maxiter$ iterations at every position.
Under causal attention, earlier positions expose only their base stream, so a query at depth $m$ attends to $P+t-1+m$ keys; when every iteration keeps its own KV stream, it attends to $P+t$ keys.
\name-fixed uses causal attention and applies the LM head and updater as \name does.
\ouro and \huginn keep one KV stream per iteration and apply the LM head only after the final iteration; \huginn executes $7+14\maxiter+7$ of the 28 layers and its input adapter once per core iteration.
This gives
\begin{align}
    \decflops^{\text{\std}}&=\sum_{t=1}^{N}\big[\flops_{\mathrm{bb}}+\flops_{\mathrm{head}}+4Ld\,(P+t)\big],\\
    \decflops^{\text{\name-fixed}}&=\sum_{t=1}^{N}\Big[\maxiter\big(\flops_{\mathrm{bb}}+\flops_{\mathrm{head}}\big)+(\maxiter-1)\,\flops_{\mathrm{upd}}\nonumber\\
    &\hphantom{{}=\sum_{t=1}^{N}\Big[}+4Ld\sum_{m=1}^{\maxiter}(P+t-1+m)\Big],\\
    \decflops^{\text{\ouro}}&=\sum_{t=1}^{N}\big[\maxiter\,\flops_{\mathrm{bb}}+\flops_{\mathrm{head}}+4Ld\,\maxiter(P+t)\big],\\
    \decflops^{\text{\huginn}}&=\sum_{t=1}^{N}\Big[\tfrac{14+14\maxiter}{28}\big(\flops_{\mathrm{bb}}+4Ld\,(P+t)\big)\nonumber\\
    &\hphantom{{}=\sum_{t=1}^{N}\Big[}+\flops_{\mathrm{head}}+2\maxiter d^2\Big].
\end{align}

\subsubsection{Training FLOPs}
\label{sec:appendix/flops/train}
\xhdr{Compact form}
Approximating backward cost as twice the corresponding forward cost,
\begin{equation}
    \trainflops\approx
    3\,\flops_{\mathrm{forward}}
    +\flops_{\mathrm{label}},
    \label{eq:trainflops}
\end{equation}
where $\flops_{\mathrm{forward}}$ includes all gradient-tracked forward computation, including the LM head, updater and decider, and $\flops_{\mathrm{label}}$ includes all additional no-gradient computation for online labels, including context reconstruction and auxiliary-module calls.
All models share the same training implementation, so implementation-level choices such as activation checkpointing are not counted and the comparison reflects only algorithmic differences.

\xhdr{Per-depth counts}
Consider a training sequence of $T$ tokens.
Let $N_m=|\{t:m_t\ge m\}|$ be the number of tokens that execute iteration $m$, with $N_1=T$, and let $A_m=\sum_{t:m_t\ge m}S_t^{(m)}$ be the number of query--key pairs at depth $m$.
Here $S_t^{(m)}$ follows Equation~\ref{eq:duo_keys} without the prompt term, because training attends over the whole sequence.

\xhdr{Gradient-tracked forward}
Every executed iteration runs the backbone and LM head with gradients, the updater prepares iterations $2$ to $\maxiter$, and the decider runs after iterations $1$ to $\maxiter-1$:
\begin{equation}
    \begin{aligned}
    \flops_{\mathrm{forward}}
    ={}&\sum_{m=1}^{\maxiter}\Big[N_m\big(\flops_{\mathrm{bb}}+\flops_{\mathrm{head}}\big)+4Ld\,A_m\Big]\\
    &+\sum_{m=2}^{\maxiter}N_m\,\flops_{\mathrm{upd}}
    +\sum_{m=1}^{\maxiter-1}N_m\,\flops_{\mathrm{decider}}.
    \end{aligned}
    \label{eq:train_forward}
\end{equation}
The decider loss reuses the decider outputs of this pass and adds no further calls.

\xhdr{Label computation}
Online labels require the next-iteration loss of every token that stops.
After iteration $m<\maxiter$, a no-gradient lookahead therefore applies the updater, backbone and LM head at depth $m+1$.
Under extended duo-causal attention, a stopped token's lookahead query must also see the depth-$(m+1)$ KV of earlier tokens that continue.
The lookahead reconstructs this context by re-running all $N_m$ tokens that executed iteration $m$, rather than only the $N_m-N_{m+1}$ tokens that stop:
\begin{equation}
    \begin{aligned}
    \flops_{\mathrm{label}}&=\sum_{m=1}^{\maxiter-1}\Big[N_m\big(\flops_{\mathrm{upd}}+\flops_{\mathrm{bb}}+\flops_{\mathrm{head}}\big)+4Ld\,\tilde A_{m+1}\Big],\\
    \tilde A_{m+1}&=\sum_{t:m_t\ge m}\Big[\sum_{u<t}\min(m_u,m+1)+m+1\Big],
    \end{aligned}
\end{equation}
where $\tilde A_{m+1}$ counts each lookahead query as if its token continued.

\xhdr{Baselines}
Baselines compute no labels, so $\flops_{\mathrm{label}}=0$, and every token executes all iterations, so $N_m=T$.
\std has one backbone pass and one LM head per token.
\ouro applies the LM head at every iteration because its objective supervises each exit, \huginn applies it once after the coda, and \name-fixed applies it at every iteration together with $\maxiter-1$ updater calls.
Their attention pairs follow the fixed-depth key counts of Section~\ref{sec:appendix/flops/decode} with $P=0$.

\xhdr{Training cost of the 1.7B runs}
Table~\ref{tab:train_flops} evaluates these expressions for all 1.7B models, each trained for 6{,}402 steps of 128 sequences (three epochs).
For \name, we use the average iteration count per token logged during training.
The label lookahead accounts for $20$--$24\%$ of the training FLOPs of \name.
At $\maxiter=2$, \name costs $1.87\times$ \std, below \name-fixed ($2.01\times$) and \ouro ($2.00\times$) at the same ceiling.

\begin{table}[!htbp]
    \caption{Training FLOPs ($10^{18}$) of the 1.7B post-training runs, split into the terms of Equation~\ref{eq:trainflops}. Fwd.\,+\,bwd.\ reports $3\,\flops_{\mathrm{forward}}$.}
    \label{tab:train_flops}
    \centering
    \fontsize{8}{10}\selectfont
    \setlength\tabcolsep{6pt}
    \begin{tabular}{llrrrr}
    \toprule
    \textbf{Model} & $\maxiter$ & \textbf{Fwd.\,+\,bwd.} & \textbf{Label} & \textbf{Total} & \textbf{vs.\ \std} \\
    \midrule
    \std & 1 & 41.8 & -- & 41.8 & 1.00$\times$ \\
    \midrule
    \multirow{2}{*}{\huginn} & 3 & 77.7 & -- & 77.7 & 1.86$\times$ \\
     & 7 & 149.3 & -- & 149.3 & 3.57$\times$ \\
    \midrule
    \multirow{2}{*}{\ouro} & 2 & 83.6 & -- & 83.6 & 2.00$\times$ \\
     & 4 & 167.1 & -- & 167.1 & 4.00$\times$ \\
    \midrule
    \multirow{2}{*}{\name-fixed} & 2 & 84.0 & -- & 84.0 & 2.01$\times$ \\
     & 4 & 168.4 & -- & 168.4 & 4.03$\times$ \\
    \midrule
    \rowcolor{tahgray}
    \cellcolor{white} & 2 & 62.8 & 15.2 & 78.1 & 1.87$\times$ \\
    \rowcolor{tahgray}
    \cellcolor{white} & 4 & 87.6 & 26.7 & 114.3 & 2.73$\times$ \\
    \rowcolor{tahgray}
    \cellcolor{white}\multirow{-3}{*}{\name} & 8 & 164.2 & 52.0 & 216.2 & 5.17$\times$ \\
    \bottomrule
    \end{tabular}
\end{table}

\section{Additional Experimental Results}
\label{sec:appendix/results}

\subsection{Teacher Selection}
\label{sec:appendix/results/teacher}
We compare Qwen3-8B, Qwen3-32B and Qwen3-235B-A22B as teachers, using responses generated from the same prompts.
We fine-tune Qwen3-1.7B-Base (\std) on each response set with an identical training recipe and evaluate with a 32K output limit, temperature $0.6$, top-$p$ $0.95$ and top-$k$ $20$.
The 8B teacher yields the highest mean student accuracy on AIME24--26 (Table~\ref{tab:teacher}), exceeding the 32B and 235B teachers by $1.63$ and $1.80$ points, respectively.
We therefore use Qwen3-8B as the teacher for all main experiments.

\begin{table}[!htbp]
    \caption{AIME accuracy (\%) for Qwen3-1.7B-Base students trained on responses from different teachers.
    All evaluations use avg@32 and a 32K output-token limit.}
    \label{tab:teacher}
    \centering
    \fontsize{8}{10}\selectfont
    \setlength\tabcolsep{6pt}
    \begin{tabular}{lcccc}
    \toprule
    \textbf{Teacher} & \textbf{AIME24} & \textbf{AIME25} & \textbf{AIME26} & \textbf{Mean} \\
    \midrule
    \rowcolor{tahgray}
    Qwen3-8B & \textbf{11.04} & \textbf{13.02} & \textbf{11.87} & \textbf{11.98} \\
    Qwen3-32B & 9.69 & 11.98 & 9.38 & 10.35 \\
    Qwen3-235B-A22B & 10.42 & 9.90 & 10.21 & 10.18 \\
    \bottomrule
    \end{tabular}
\end{table}

\subsection{Inference Depth Outside the Training Ceiling}
\label{sec:appendix/results/ood_depth}
We evaluate the $\maxiter=8$ checkpoint with the inference ceiling lowered to $4$ or raised to $12$, without retraining.
Table~\ref{tab:ood_depth} reports AIME24--26 accuracy and mean realized depth.
Lowering the ceiling to $4$ reduces accuracy by $2.7$--$3.1$ points across AIME24--26.
Raising it to $12$ leaves accuracy essentially unchanged while increasing mean depth by $20$--$27\%$.
Additional depth at inference therefore neither breaks the model nor improves it beyond the trained ceiling.

\begin{table}[!htbp]
    \caption{Effects of changing the inference depth ceiling. $M$ denotes the training ceiling; the upper group provides reference models evaluated at their training ceilings. Accuracy (\%) uses avg@32; Avg. is the mean across AIME24--26.}
    \label{tab:ood_depth}
    \centering
    \fontsize{8}{10}\selectfont
    \setlength\tabcolsep{4pt}
    \begin{tabular}{lcccccc}
    \toprule
    \textbf{Model} & \makecell{\textbf{Inference}\\\textbf{ceiling}} & \textbf{AIME24} & \textbf{AIME25} & \textbf{AIME26} & \textbf{Avg.} & \makecell{\textbf{Mean}\\\textbf{depth}} \\
    \midrule
    \std & 1 & 11.0 & 13.0 & 11.9 & 12.0 & 1.00 \\
    \name ($M=2$) & 2 & 15.7 & 16.5 & 14.3 & 15.5 & 1.23 \\
    \name ($M=4$) & 4 & 16.3 & 16.9 & 14.2 & 15.8 & 1.49 \\
    \midrule
    \name ($M=8$) & 4 & 13.6 & 15.1 & 11.4 & 13.4 & 1.55 \\
    \rowcolor{tahgray}
    \name ($M=8$) & 8 & 16.3 & 17.9 & 14.5 & 16.2 & 2.22 \\
    \name ($M=8$) & 12 & 16.9 & 17.2 & 14.5 & 16.2 & 2.72 \\
    \bottomrule
    \end{tabular}
\end{table}

\section{Additional Analysis}
\label{sec:appendix/analysis}

\subsection{Validation Loss During Post-training}
\label{sec:appendix/training/curves}
\begin{figure}[!htbp]
    \centering
    \includegraphics[width=\linewidth]{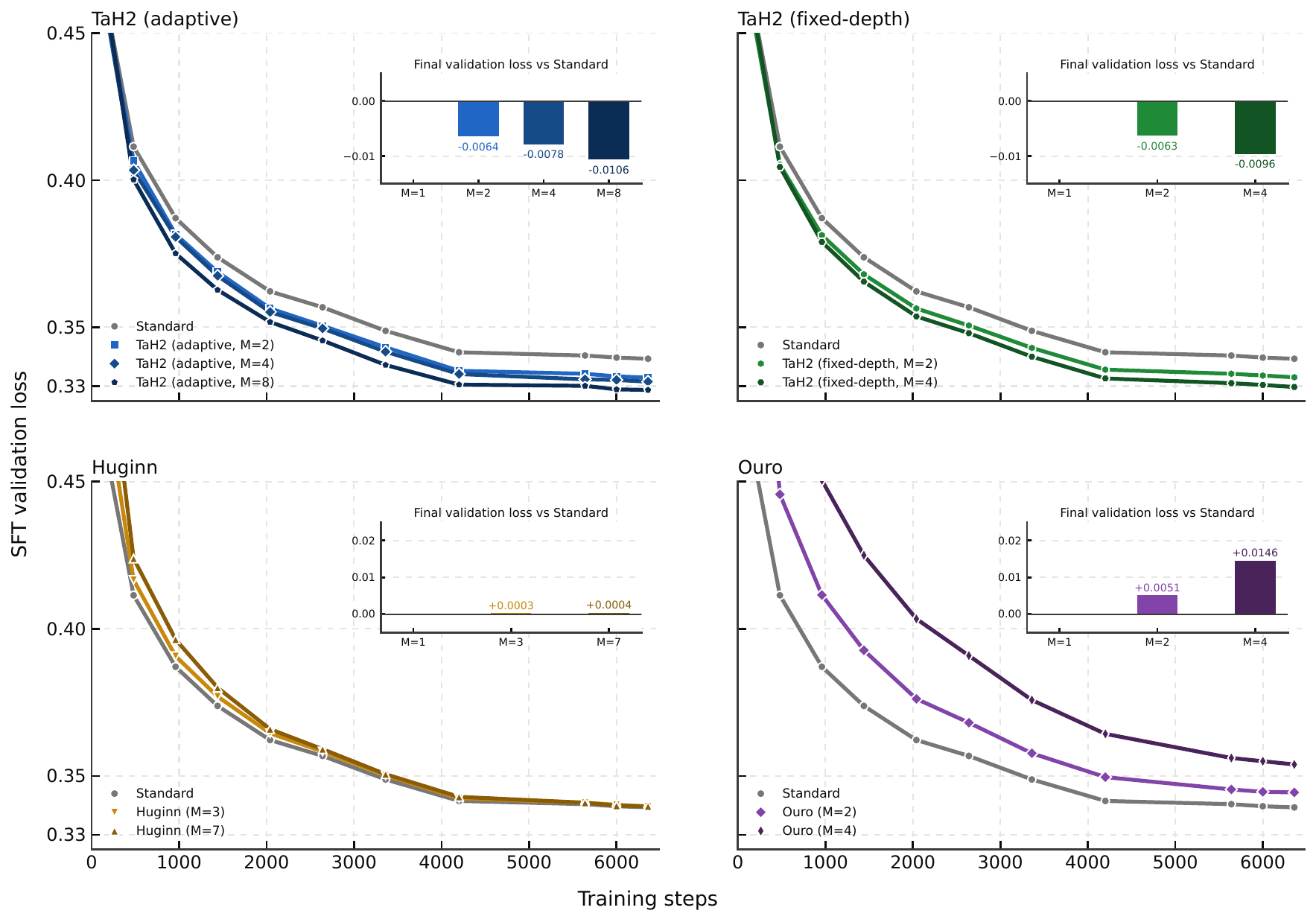}
    \caption{Validation loss during post-training at 1.7B. Insets show final loss differences from \std; negative values indicate improvements.}
    \label{fig:loss_steps}
\end{figure}

Figure~\ref{fig:loss_steps} shows validation loss throughout post-training at 1.7B.
Both \name variants maintain lower loss than \std, with greater iteration depth yielding further reductions and adaptive \name attaining the lowest final loss.

\subsection{Per-Benchmark Test-Time Scaling}
\label{sec:appendix/results/tts}

\begin{figure*}[!htbp]
    \centering
    \includegraphics[width=0.98\linewidth]{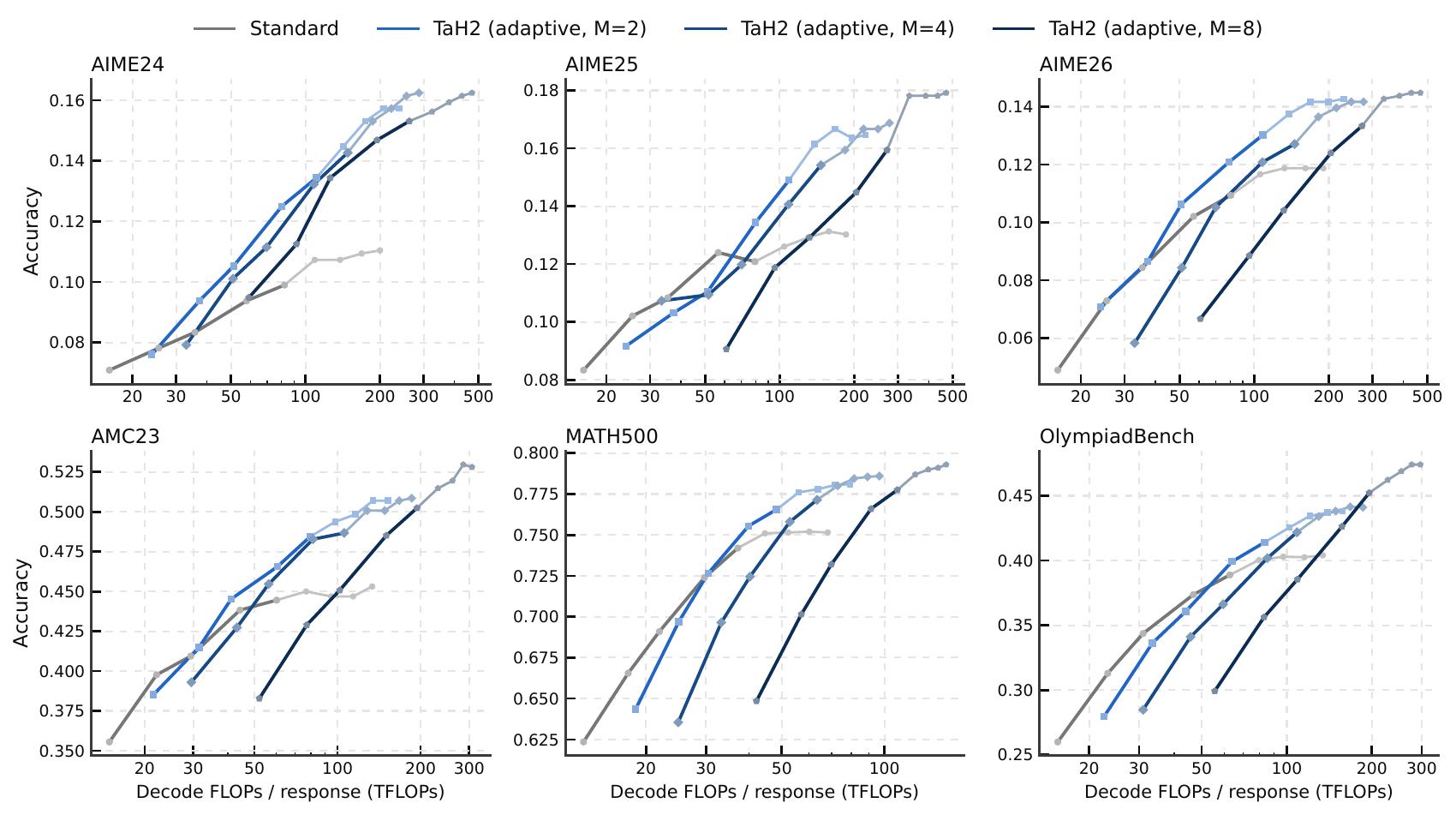}
    \caption{Per-benchmark test-time scaling on the six math benchmarks at 1.7B.
    Each panel compares \std with \name at $\maxiter\in\{2,4,8\}$ using the same nine output-token cutoffs as Figure~\ref{fig:acc_tts}.
    Dark segments extend through 16K and light segments through 32K.}
    \label{fig:depth/tts_benchmarks}
\end{figure*}

Figure~\ref{fig:depth/tts_benchmarks} shows test-time scaling on each of the six math benchmarks.
We use output-token cutoffs of 4K, 6K, 8K, 12K, 16K, 20K, 24K, 28K and 32K, matching the sampling grid of Figure~\ref{fig:acc_tts}.

\subsection{Parallel Test-Time Scaling}
\label{sec:appendix/results/cons}
\begin{figure}[!htbp]
    \centering
    \includegraphics[width=\linewidth]{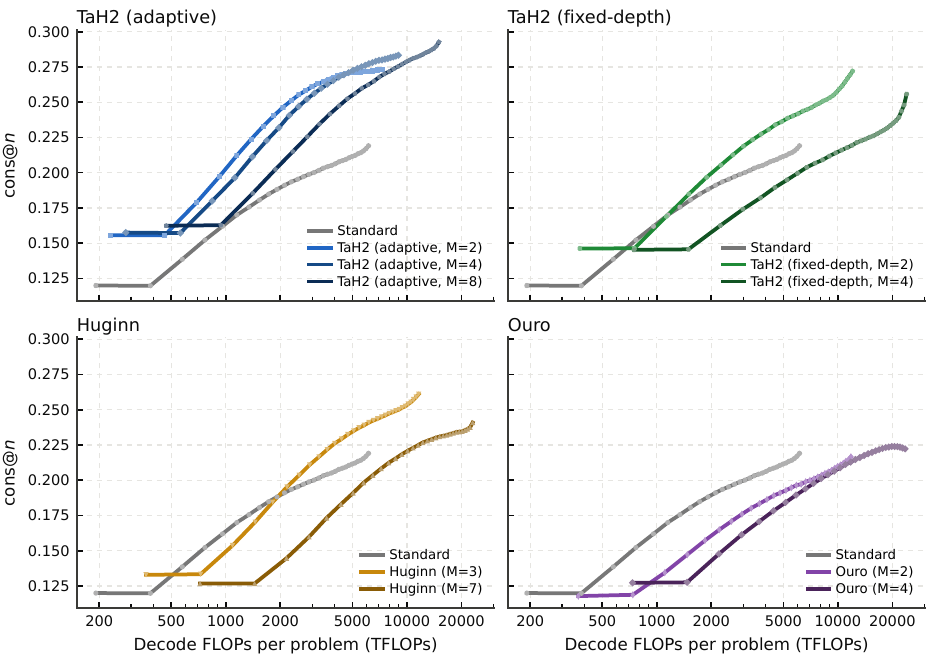}
    \caption{Parallel test-time scaling: mean AIME24--26 cons@$n$ versus decoding FLOPs per problem for $n=1,\dots,32$, one model family per panel with \std for reference.}
    \label{fig:cons_at_n}
\end{figure}

Test-time compute can also scale in parallel, by sampling several responses and aggregating them with majority voting.
Figure~\ref{fig:cons_at_n} measures this axis with cons@$n$ on AIME24--26: for each problem, we vote over $n$ of the 32 recorded full-length responses, drawn without replacement and averaged over random draws, and count decoding FLOPs as $n$ times the mean per-response cost.
\name at $\maxiter=2$ reaches $27.3\%$ at $n=32$ compared with $21.9\%$ for \std, and its curve stays above every fixed-depth baseline at matched FLOPs.

\noindent
\subsection{Token-Level Depth Allocation}
\label{sec:appendix/analysis/token_depth}
\begin{figure}[!htbp]
    \centering
    \includegraphics[width=\linewidth]{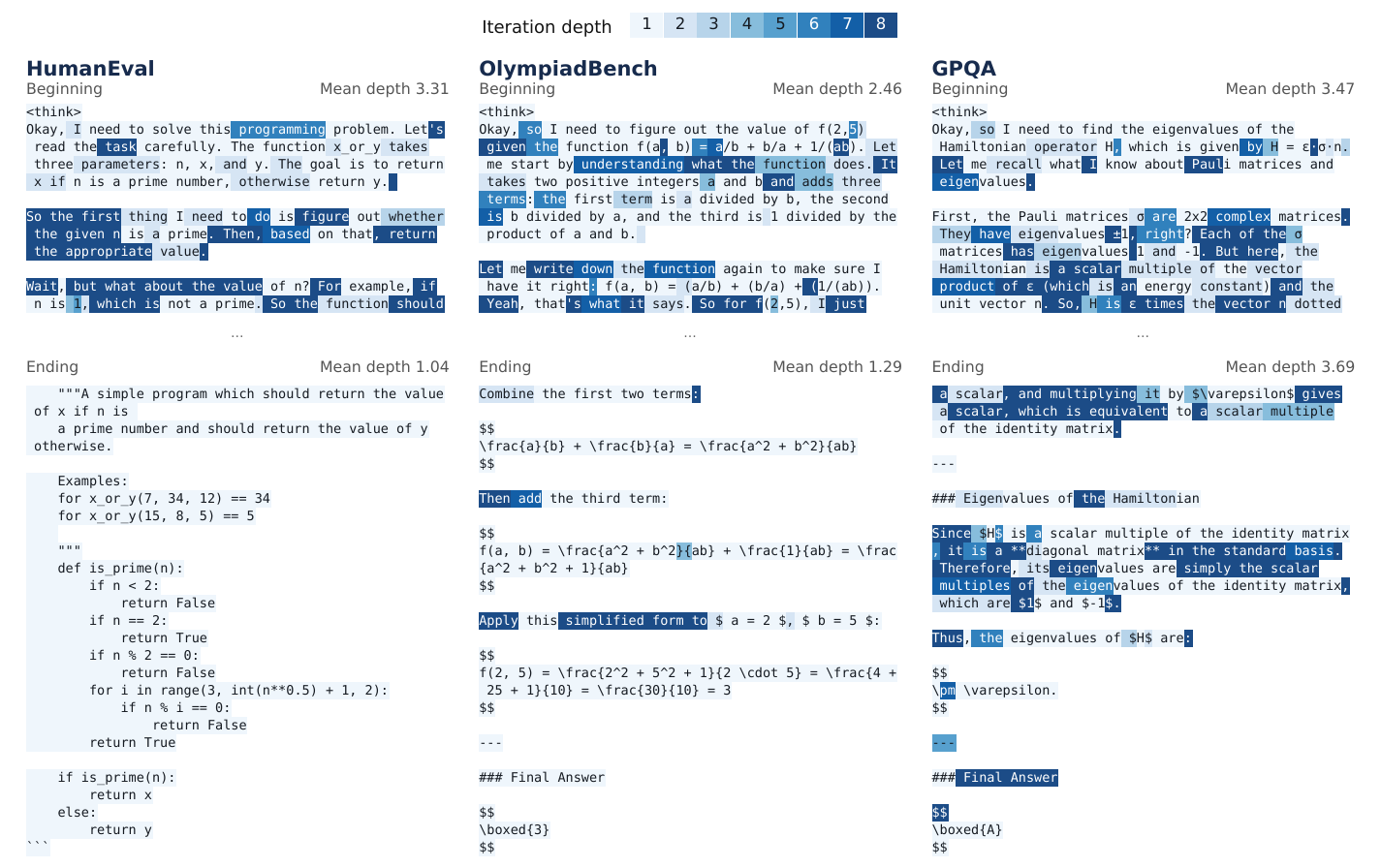}
    \caption{Token-level iteration depth in sampled correct responses from OlympiadBench, HumanEval and GPQA. Darker shading indicates more iterations; mean depths refer to the displayed spans.}
    \label{fig:depth_vs_token}
\end{figure}

Figure~\ref{fig:depth_vs_token} shows iteration depths in three sampled correct responses.
The math and code examples show a clear shift from deeper natural-language reasoning to shallower mathematical expressions and final code.
Mean depth falls from $2.46$ to $1.29$ in OlympiadBench and from $3.31$ to $1.04$ in HumanEval across the displayed spans.
The GPQA example retains substantial depth through its concluding explanation.
These cases suggest that depth allocation reflects local response content and the stage of reasoning.

\subsection{Cross-Iteration Attention}
\label{sec:appendix/analysis/attention}
\begin{figure}[!htbp]
    \centering
    \includegraphics[width=\linewidth]{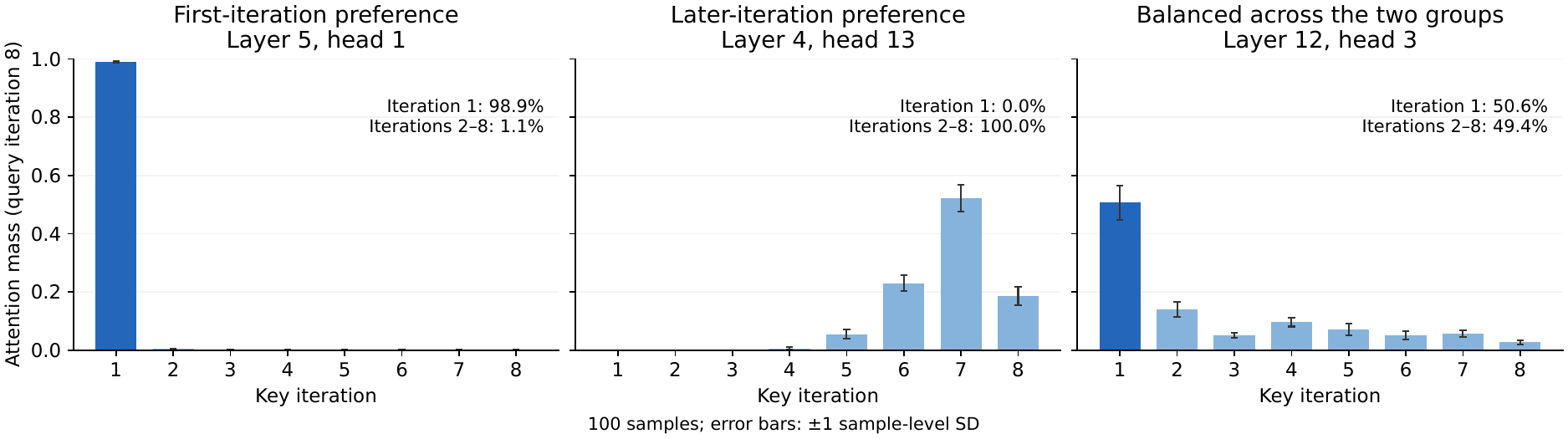}
    \caption{Attention mass by key iteration for three representative heads at query iteration 8. Bars show means over 100 sequences; error bars indicate one sample-level standard deviation. Queries are averaged within each sequence first.}
    \label{fig:appendix/attn_prefer}
\end{figure}

\begin{figure}[!htbp]
    \centering
    \includegraphics[width=\linewidth,height=0.84\textheight,keepaspectratio]{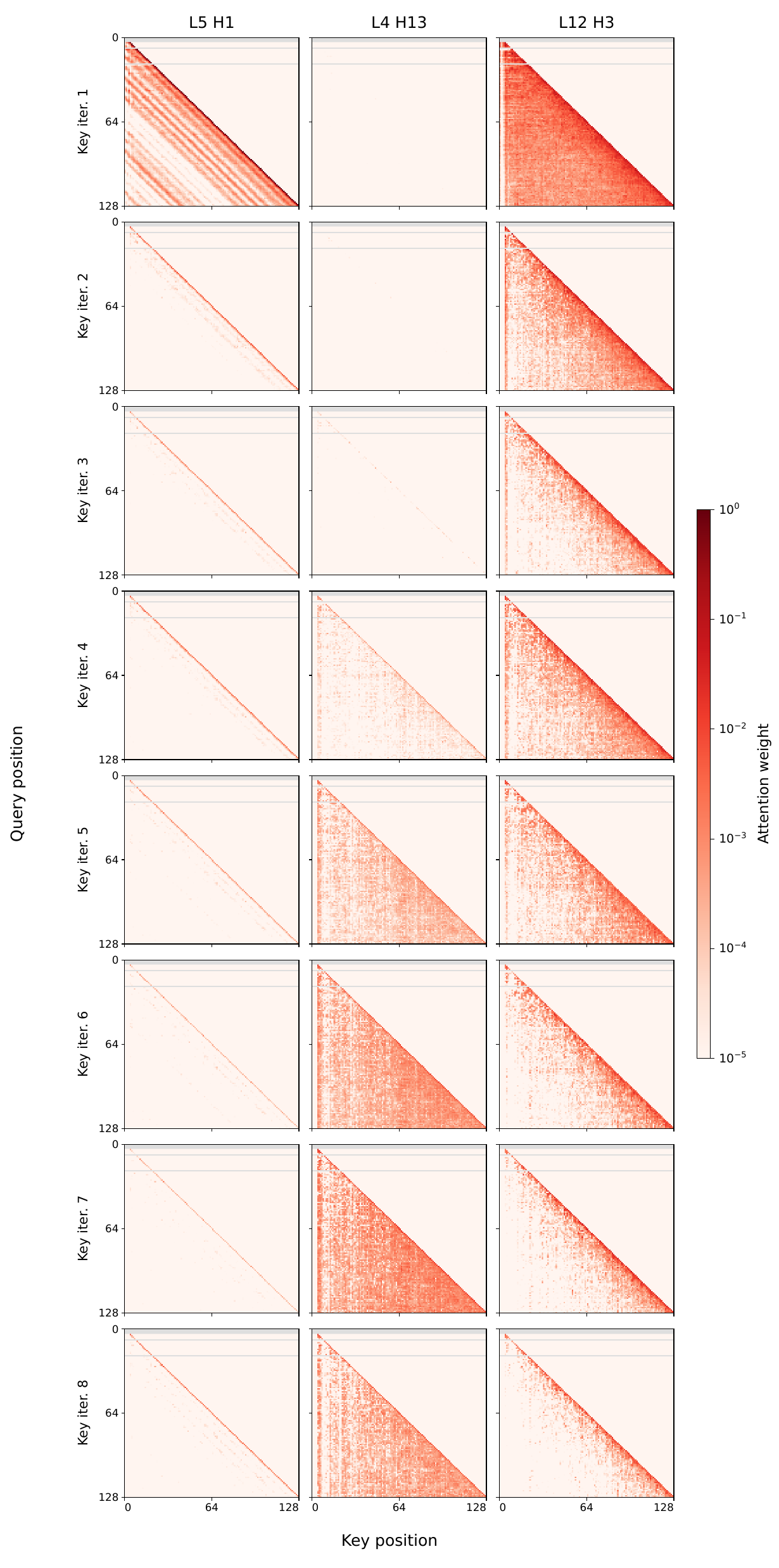}
    \caption{Attention maps for the same heads, with columns indexing heads and rows indexing key iterations. Each query position is averaged over sequences that reach iteration 8. All panels share a logarithmic colour scale; grey marks positions with no executed queries. Prompt keys are omitted without renormalising attention.}
    \label{fig:appendix/attn_split}
\end{figure}

We examine three representative heads of \name ($\maxiter=8$) on 100 randomly sampled validation sequences under teacher forcing, retaining the decider's actual routes.
For queries at iteration 8, Figure~\ref{fig:appendix/attn_prefer} shows distinct preferences for first-iteration states, later-iteration states, or both.
Figure~\ref{fig:appendix/attn_split} visualises these patterns over the first 128 response positions with the original context preserved.

\section{Full Formulation of \Supervision}
\label{sec:appendix/formulation}

\xhdr{Continuation labels and gain coverage}
The continue target $c_t^{(m)}$ indicates whether token $t$ should execute another iteration, whereas $m_t$ records how many iterations token $t$ executes under the current decider.
As in Section~\ref{sec:method/sft}, $a_t^{(m)}=\1[m_t\ge m]$ records whether supervised token $t$ executes iteration $m$.
These labels are recomputed from the current backbone on every training batch and do not determine the routes used in that forward pass.
We initialise $c_t^{(0)}=1$ for every supervised token and compute the loss reduction $\delta_t^{(m)}=\ell_t^{(m)}-\ell_t^{(m+1)}$ as in Equation~\ref{eq:iteration_gain}.
For a token that stops, a no-gradient lookahead supplies its next-iteration loss.

At iteration $m$, collect the positive gains of tokens with $a_t^{(m)}=1$ and $c_t^{(m-1)}=1$ across data-parallel ranks and sort them as $\delta_{[1]}^{(m)}\ge\dots\ge\delta_{[n]}^{(m)}>0$, where $[i]$ denotes gain rank and $n$ is the number of positive gains.
The coverage cutoff is
\begin{equation}
    i_{\mathrm{cov}}=\min\Big\{i:\sum_{i^{\prime}=1}^{i}\delta_{[i^{\prime}]}^{(m)}\ge\rho\sum_{i^{\prime}=1}^{n}\delta_{[i^{\prime}]}^{(m)}\Big\},
    \qquad
    \delta_{\mathrm{cut}}^{(m)}=\delta_{[i_{\mathrm{cov}}]}^{(m)}.
\end{equation}
Equation~\ref{eq:labels} assigns continue labels to gains at or above this cutoff, including ties.
Thus $\rho$ specifies a fraction of total positive gain, not a fraction of tokens.
As an exception to Equation~\ref{eq:labels}, if there are no positive gains, we set $\delta_{\mathrm{cut}}^{(m)}=0$ only for weight computation and assign zero continue labels at this and all later iterations.

\xhdr{Cost-sensitive weights}
For tokens with $a_t^{(m)}=1$, the per-token weights are
\begin{equation}
    w_t^{(m)}=
    \begin{cases}
        \operatorname{clip}\!\left(|\delta_t^{(m)}-\delta_{\mathrm{cut}}^{(m)}|,\,10^{-6},\,1\right), & c_t^{(m-1)}=1,\\[2pt]
        10^{-6}, & c_t^{(m-1)}=0.
    \end{cases}
\end{equation}
The fixed lower bound preserves a small supervision weight near the cutoff and for tokens already labelled to stop; the upper bound limits the influence of large loss changes.
Labels, cutoffs and weights are treated as constants during backpropagation.

\xhdr{Class balancing and joint loss}
At each iteration, we count stop and continue labels among tokens with $a_t^{(m)}=1$, pooling counts across data-parallel ranks.
The positive-class balancing weight $\beta^{(m)}$ is their stop-to-continue count ratio; it is set to one if either class is absent.
The BCE in Equation~\ref{eq:loss} is therefore
\begin{equation}
    \mathrm{BCE}\!\left(g_t^{(m)},c_t^{(m)}\right)
    =-\beta^{(m)}c_t^{(m)}\log g_t^{(m)}
     -(1-c_t^{(m)})\log(1-g_t^{(m)}).
\end{equation}
The coverage cutoff and class-balancing weights are calculated from each batch, with the clipping bounds above fixed throughout the experiments.

\section{Additional Related Work}
\label{sec:appendix/related}
This section extends Section~\ref{sec:related_work}, grouped by test-time compute, recurrent design, adaptive depth and serving.

\xhdr{Test-time scaling}
Token-based scaling further differs in how it controls reasoning length and selects candidates.
Within a sequence, online confidence adjusts reasoning budgets~\citep{li2026rebalance}, while analyses of overthinking show that longer chains are not uniformly better~\citep{wu2025morelesscot}.
Across candidates, process-reward search~\citep{guan2025rstar}, likelihood-based sampling~\citep{karan2025sampling} and effort-aware sample scheduling~\citep{chen2026thinkdeep} guide exploration; call-count studies examine the non-monotone returns of additional LLM calls~\citep{chen2024compound}.
Within latent representations, methods compress multimodal reasoning~\citep{shen2025heima}, supervise parallel latent blocks~\citep{fan2026lotus}, switch between latent and discrete reasoning by confidence~\citep{xu2026thinkrouter}, or add hidden sequence positions~\citep{liu2026hiddendecoding}.
ParScale instead adds parallel computation through multiple streams sharing model parameters~\citep{chen2025parscale}.

\xhdr{Recurrent architectures}
Recurrent architectures differ in what they repeat and retain.
Universal Transformers share transformations across depth~\citep{dehghani2018universal}, whereas block-recurrent Transformers carry recurrent state across sequence blocks~\citep{hutchins2022block}.
Within depth recurrence, studies of growing and looping examine which blocks to repeat~\citep{kapl2026growinglooping}; Loopies repeats individual layers, offering an alternative to middle-block and full-stack recurrence~\citep{gao2026loopies}.
Pondering models feed predictions back as embeddings~\citep{zeng2025pondering}, CoTFormer interleaves intermediate representations as additional tokens~\citep{mohtashami2023cotformer}, and recursive reasoners refine latent states on structured tasks~\citep{ge2025hrmperspectives,ren2026hrmguessing,baek2026gram,altabaa2025recursivelatent}.
Theory examines the expressivity and memory requirements of repeated computation relative to explicit CoT~\citep{saunshi2025loopedTrans,zhang2026memorybudget}.

\xhdr{Scaling under resource constraints}
Scaling results depend on which resources are held fixed as recurrence increases.
Parcae studies compute allocation at fixed unique parameter count, allowing effective depth and per-token computation to grow~\citep{prairie2026parcae}.
Iso-Depth instead fixes effective depth and quantifies the capacity retained when independent blocks are replaced by shared iterations~\citep{schwethelm2026isodepth}.
Sparse-layer studies examine how expert routing affects the cost of parameter sharing~\citep{lee2026sparselayers}, and Loopies compares layer-looped MoE models under matched wall-clock training budgets~\citep{gao2026loopies}.
SMELT matches per-token FLOPs, total non-embedding parameters and KV cache, then fits separate pretraining scaling laws for looped and non-looped MoE models~\citep{wang2026smelt}.
Architectural syntheses likewise distinguish parameter efficiency from computational cost~\citep{huang2026loopedmodels}.
Our comparison varies output-token budgets after post-training and measures reasoning accuracy against decoding FLOPs per response, complementing these architectural and pretraining studies.

\xhdr{Depth scaling and stability}
Increasing the depth used during training and executing more iterations than were seen during training are distinct settings.
\ouro, Parcae and STARS report saturation or degradation when inference extends beyond the trained depth regime~\citep{zhu2025ouro,prairie2026parcae,yang2026stars}; RecurTrace, \tah and code-model studies also document non-monotone returns from additional iterations~\citep{wang2026recurtrace,fu2025tah,yang2026loopcoderv2}.
Stability methods, including STARS, modify residual scaling, input injection or recurrent dynamics~\citep{li2026deeploop,wang2026residualscaling,fu2026fullylooped,yang2026stars,labovich2026stability}, while readout analyses show that per-loop supervision constrains only the state variables exposed to the readout~\citep{sharma2026readout}.
Mechanistic studies distinguish state convergence from predictive improvement~\citep{blayney2026mechanistic,viakhirev2026thinkshallow}, examining two-scale dynamics~\citep{pappone2025twoscale}, dynamical regimes~\citep{kim2026phaseselection,zhang2026convergenceselection}, Jacobian structure~\citep{wang2026jacobianlens} and links between latent and verbal reasoning~\citep{chen2026loopbridge}.
Our depth-scaling experiments increase the maximum iteration depth used in training; Appendix~\ref{sec:appendix/results/ood_depth} examines inference beyond it.

\xhdr{Recurrent states and output aggregation}
State design determines which information remains available across iterations.
Memory highways~\citep{yu2025mesh} and depth attention~\citep{knupp2026dreamer} expose earlier states, while anchored injection~\citep{kim2026scse}, discrete--continuous channels~\citep{fu2026discoloop} and gated modulation~\citep{hegazy2026grt} alter state updates.
Other work changes the repeated computation through mixer-only loops~\citep{lin2026mixerloop}, multi-token guidance~\citep{shomali2026loopmtp} or denoising objectives~\citep{suleymanzade2026loopedflows}.
For output aggregation, PonderLM-3 weights hidden states by predicted depth probabilities~\citep{li2026ponderlm3}; Adaptive Latent CoT and Adaptive Loops and Memory use halting probabilities to combine intermediate states~\citep{zeng2026adaptivelatentcot,frey2026adaptiveloops}.
\name uses input injection between iterations and mixes output distributions across executed depths, with the final depth absorbing the remaining stopping mass.

\xhdr{Post-training recurrence into pretrained LLMs}
Conversion methods differ in how they preserve and adapt pretrained computation, paralleling upcycling into MoE or deeper models~\citep{komatsuzaki2022upcycling,kim2023solar,wu2024llamapro}.
Beyond the conversions in Section~\ref{sec:related_work}, path-preserving initialisation retains single-pass behaviour~\citep{shapiro2026retrofit}, a trainable recurrent module can be attached to a frozen backbone~\citep{panwar2026relit}, and middle layers can be repeated without training~\citep{chen2026trainingfree}.
Retrofitted recurrence has been evaluated on math and compositional tool use~\citep{mcleish2025retrofit,popescu2026toolcalling}.
DND learns selective recomputation with routing regularisation and target selection ratios~\citep{chen2025dnd}; \tah trains a decider from offline mismatch labels in a separate stage~\citep{fu2025tah}.
\name instead jointly adapts the backbone and decider using online supervision of iteration gains.

\xhdr{Dimensions of adaptive computation}
Adaptive architectures select computation within layers or across depth.
Along width, sparse MoE routing selects experts for each token~\citep{fedus2022switch}; ZEDA further varies expert activation through zero-expert injection and self-distillation~\citep{lv2026zeda}.
Its token-level visualisations show reduced computation for mathematical expressions and code fragments relative to natural-language reasoning, relating compute allocation to response content.
Along depth, LayerSkip enables early exits~\citep{elhoushi2024layerskip}, while MoD routes selected tokens through each layer~\citep{raposo2024mixture}.
Other approaches change computation by routing to larger models~\citep{fu2025r2r} or grouping tokens into concepts~\citep{huang2026conceptmoe,qu2025dlcm}.

\xhdr{Adaptive iteration depth}
Within looped models, depth decisions differ in granularity and timing.
LoopFormer follows a user-specified sequence budget~\citep{jeddi2026loopformer}; T-LoopFormer and PonderLM-3 predict token depths from initial states~\citep{yu2026tloopformer,li2026ponderlm3}; ANIRA compares such initial allocation with decisions after each iteration~\citep{moosa2026anira}.
Training signals provide a separate distinction.
AdaPonderLM trains iterative gates with a compute penalty~\citep{song2026adaponderlm}, whereas RL-Halting uses terminal rewards to learn sequence-level stopping~\citep{kuo2026rlhalting}.
Explicit gain supervision is also used by \ouro's second-stage gate and RecurTrace's separately trained sequence-level head~\citep{zhu2025ouro,wang2026recurtrace}; \name derives token-level targets online from the evolving backbone during joint training.
Alternatives use a learned confidence head~\citep{park2026loopus} or convergence-based stopping rules~\citep{geiping2025scaling,logan2026pertoken,movahedi2026fprm,pappone2025twoscale}.
Gate collapse has been reported for \ouro, RecurTrace and retrofitted models~\citep{zhu2025ouro,wang2026recurtrace,shapiro2026retrofit} under particular objectives and training settings, and we observe it for \ouro's Stage I in our post-training setting (Appendix~\ref{sec:appendix/arch/definitions}).
Controlled diagnoses examine how jointly learned gates affect training trajectories~\citep{popescu2026adaptivedepth}, while analyses of hierarchical recurrent models find settings where fixed maximum depth outperforms adaptive halting~\citep{ge2025hrmperspectives}.

\xhdr{Serving adaptive depth}
Whether FLOPs savings become wall-clock gains depends on execution.
Continuous depth batching schedules tokens at different depths in shared passes~\citep{bae2024relaxed,schwethelm2026cdb}, cross-loop parallelism and diffusion-style samplers overlap iterations across tokens~\citep{wu2025plt,geiping2025parallelsamplers}, and cross-loop KV sharing or compression bounds memory~\citep{vendrell2026melt,deng2026lt2,oneill2026loopedlatentattn}.
CHASE adapts training to the missing states left by skipped iterations under early exit~\citep{yu2026chase}.
We report decoding FLOPs as the primary metric and verify throughput with a depth-batched engine (Section~\ref{sec:exp/performance}).

\end{document}